\documentclass[pmlr]{jmlr}% new name PMLR (Proceedings of Machine Learning)

\usepackage{longtable}% for long tables

 \usepackage{booktabs}
 
\usepackage[load-configurations=version-1]{siunitx} % newer version
\usepackage{url}
\usepackage{amsfonts,bm,dsfont}
\usepackage{makecell}
\usepackage{hhline}
\usepackage{xcolor}
\usepackage{colortbl}
\usepackage[normalem]{ulem}
\usepackage{caption}
\usepackage{enumitem}
\usepackage{layouts}

\usepackage{pgfplotstable}
\usepackage{pgfplots}

\makeatletter
\def\set@curr@file#1{\def\@curr@file{#1}} %temp workaround for 2019 latex release
\makeatother

\newcommand\sL{\ensuremath{\mathcal{L}}}

\newcommand\sT{\ensuremath{\mathcal{T}}}

\newcommand\bh{\ensuremath{\mathbf{h}}}

\newcommand\bu{\ensuremath{\mathbf{u}}}
\newcommand\bv{\ensuremath{\mathbf{v}}}

\newcommand\bx{\ensuremath{\mathbf{x}}}

\newcommand\bW{\ensuremath{\mathbf{W}}}

\newcommand\BR{\ensuremath{\mathbb{R}}}

\newcommand\R{\ensuremath{\mathbb{R}}} % Real numbers
\newcommand\refsec[1]{Section~\ref{sec:#1}}

\newcommand\reffig[1]{Figure~\ref{fig:#1}}

\newcommand\reftab[1]{Table~\ref{tab:#1}}
\newcommand\refapp[1]{Appendix~\ref{sec:#1}}

\usepackage{color}

\definecolor{lblue}{HTML}{A6CEE3}
\definecolor{lgreen}{HTML}{B2DF8A}
\definecolor{lred}{HTML}{FB9A99}
\definecolor{lorange}{HTML}{FDBF6F}
\definecolor{mblue}{HTML}{80B1D3}
\definecolor{mgreen}{HTML}{B3DE69}
\definecolor{mred}{HTML}{FB8072}
\definecolor{morange}{HTML}{FDB462}
\definecolor{blue}{HTML}{1F78B4}
\definecolor{green}{HTML}{33A02C}
\definecolor{red}{HTML}{E31A1C}
\definecolor{orange}{HTML}{FF7F00}
\definecolor{dblue}{HTML}{08519C}
\definecolor{dgreen}{HTML}{006D2C}
\definecolor{dorange}{HTML}{EC7014}

\newcommand{\cc}{\cellcolor[gray]{0.92}}

\newcommand{\tbf}[1]{{\textbf{#1}}}

\renewcommand{\cite}{\citep}
\usepackage{hyperref}

\makeatletter
\newcommand{\printfnsymbol}[1]{%
  \textsuperscript{\@fnsymbol{#1}}%
}
\theorembodyfont{\upshape}
\theoremheaderfont{\scshape}
\theorempostheader{:}
\theoremsep{\newline}

\jmlrvolume{182}
\jmlryear{2022}
\jmlrworkshop{Machine Learning for Healthcare}

\title[Contrastive Learning of Medical Visual Representations from Paired Images and Text]{Contrastive Learning of Medical Visual Representations\\ from Paired Images and Text}

\author{\Name{Yuhao Zhang}\thanks{The first two authors contributed equally. YZ is now affliated with AWS AI Labs, while the work was done before his current affiliation. HJ is now affliated with Massachusetts Institute of Technology.}
      \Email{yuhao@cs.stanford.edu}\\ 
      \addr Biomedical Informatics Training Program, Stanford University\\
      \AND
      \Name{Hang Jiang}\printfnsymbol{1}
      \Email{hjian42@stanford.edu}\\ 
      \addr Symbolic Systems Program, Stanford University\\
     \AND
      \Name{Yasuhide Miura}\thanks{YM is now affiliated with FUJIFILM Corporation.}
      \Email{ysmiura@stanford.edu}\\ 
      \addr Computer Science Department, Stanford University\\
     \AND
      \Name{Christopher D. Manning}
      \Email{manning@stanford.edu}\\ 
      \addr Computer Science and Linguistics Departments, Stanford University\\
     \AND
      \Name{Curtis P. Langlotz}
      \Email{langlotz@stanford.edu}\\ 
      \addr Department of Radiology, Stanford University\\
} 

\begin{document}

\maketitle

\begin{abstract}
  Learning visual representations of medical images (e.g., X-rays) is core to medical image understanding but its progress has been held back by the scarcity of human annotations.
Existing work commonly relies on fine-tuning weights transferred from ImageNet pretraining, which is suboptimal due to drastically different image characteristics,
or rule-based label extraction from the textual report data paired with medical images, which is inaccurate and hard to generalize.
Meanwhile, several recent studies show exciting results from unsupervised contrastive learning from natural images, but we find these methods help little on medical images because of their high inter-class similarity.
We propose ConVIRT, an alternative unsupervised strategy to learn medical visual representations by exploiting naturally occurring paired descriptive text.
Our new method of pretraining medical image encoders with the paired text data via a bidirectional contrastive objective between the two modalities is domain-agnostic, and requires no additional expert input.
We test ConVIRT by transferring our pretrained weights to 4 medical image classification tasks and 2 zero-shot retrieval tasks, and show that it leads to image representations that considerably outperform strong baselines in most settings.
Notably, in all 4 classification tasks, our method requires only 10\% as much labeled training data as an ImageNet initialized counterpart to achieve better or comparable performance, demonstrating superior data efficiency.
\end{abstract}

\section{Introduction}
\label{sec:intro}

Medical image understanding has the potential to transform healthcare and has seen rapid progress with deep learning \cite{gulshan2016retinal, esteva2017dermatologist, de2018clinically, rajpurkar2018chexnext}.
Yet, with expert-level performance achieved only in some specialties and under some circumstances, medical image understanding remains a difficult task, with classifications dependent on subtle visual distinctions in overall similar images.
This is further exacerbated by the extreme scarcity of annotated data.

\begin{figure}[ht!]
% \shrinkcaptionmargin
\small
\centering
\includegraphics[width=0.5\textwidth]{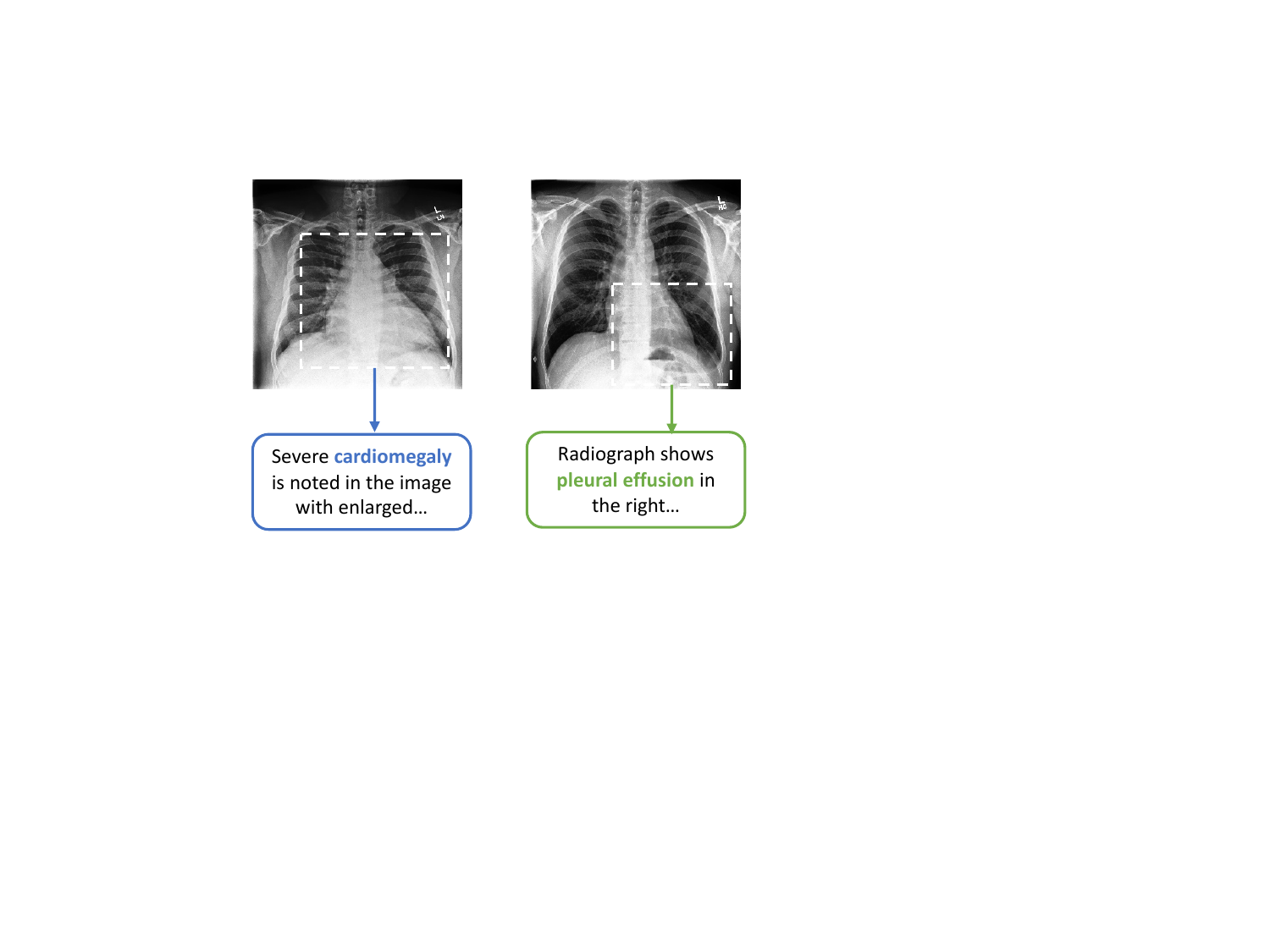}
\caption{Two example chest X-ray images with different abnormality categories, along with sentences from their paired textual report and example views indicative of their characteristics.}
% \shrinkcaptionmargin
\label{fig:intro}
\end{figure}

Existing work has followed two general approaches to obtain annotations for medical imaging tasks.
The first approach has been using high-quality annotations created by medical experts \cite{abramoff2016improved, gulshan2016retinal, shih2019augmenting, wang2020covid}.
However, the high cost of this approach has resulted in datasets that are mostly orders of magnitude smaller than natural image datasets such as ImageNet \cite{russakovsky2015imagenet}.
To remedy this, existing work has relied heavily on transferring model weights from ImageNet pretraining \cite{wang2017chestx, esteva2017dermatologist, irvin2019chexpert}.
This approach is suboptimal because, as shown in \reffig{intro}, medical image understanding often requires representations of very fine-grained visual features that are drastically different from those required for identifying objects in natural images.
As a result, \citet{raghu2019transfusion} found that ImageNet pretraining often provides little to no benefit compared to simple random initialization.

A second popular approach is to use expert-crafted rules to extract labels from the textual reports accompanying the images.
This approach has led to datasets of larger scale, since the text data paired with medical images are often produced naturally by medical experts in their routine workflow and abundant in a typical hospital's IT systems.
Nevertheless, this rule-based label extraction approach has two key limitations:
1) the rules are often inaccurate and limited to a few categories \cite{wang2017chestx}, leading to very inefficient use of the textual report data;
2) these rules are often domain-specific and sensitive to the style of the text, making cross-domain and cross-institution generalization difficult \cite{irvin2019chexpert}.

In efforts to make more efficient use of unlabeled image data, several recent studies have shown promising results from contrastive representation learning from natural images \cite{chen2020simple, he2020momentum, grill2020bootstrap}.
However, as we will show, applying these image view--based contrastive methods to medical images provides only marginal benefits compared to ImageNet pretraining, a result mostly due to the high inter-class similarity of the medical images as in \reffig{intro}.

In this work, we introduce a new method to improve visual representation learning on medical images by combining the benefits of both learning from abundant textual data and unsupervised statistical approaches.
We present \textit{\tbf{Con}trastive \tbf{VI}sual \tbf{R}epresentation Learning from \tbf{T}ext (ConVIRT)}, a framework for learning visual representations by exploiting the naturally occurring pairing of images and textual data.
ConVIRT improves visual representations by maximizing the agreement between true image-text pairs versus random pairs via a bidirectional contrastive objective between the image and text modalities.
We apply ConVIRT to the pretraining of medical image encoders, and show that it leads to higher-quality in-domain image representations that capture the subtlety of visual features required for medical image understanding tasks.

Compared to existing methods, ConVIRT has the advantages of utilizing the paired text data in a way agnostic to the medical specialty and requiring no additional expert input.
This allows us to evaluate ConVIRT by transferring our pretrained encoder weights to 4 different medical image classification tasks covering 2 medical specialties.
We find that the resulting models outperform all baseline initialization approaches, including the widely used ImageNet pretraining and strong baselines that also utilize the paired text data.
It further improves upon popular image-only unsupervised learning methods such as SimCLR \cite{chen2020simple} and MoCo v2 \cite{chen2020improved}.
Most notably, in all 4 classification tasks, ConVIRT requires only 10\% as much labeled training data as an ImageNet initialized counterpart to achieve better or comparable performance.
We further evaluate ConVIRT on two new zero-shot retrieval tasks, an image-image and a text-image retrieval task, and also find it superior to all baselines.

Since its original release in 2020, ConVIRT has directly inspired subsequent studies such as the CLIP framework \cite{radford2021learning} and the ALIGN model \cite{jia2021scaling}, which showed that direct adaptations of ConVIRT-style pretraining at much larger scales lead to state-of-the-art general visual recognition capabilities. To facilitate future research, we make our model and the collected retrieval datasets\footnote{\url{https://github.com/yuhaozhang/convirt}} publicly available.

\subsection{Generalizable Insights about Machine Learning in the Context of Healthcare}
Healthcare data is usually scarce and costly to annotate compared to data in the general domain.
As a result, machine learning models built with a single modality of healthcare data often face the generalization challenge due to small sample sizes of training data.
Meanwhile, healthcare data is often naturally paired with multimodal clinical features, including text descriptions or patient metadata, which can be exploited to reduce the cost of building reliable machine learning models.
Our method, ConVIRT, demonstrates an application of this idea to learning robust medical image encoders by reusing descriptive text naturally produced by experts via a cross-modality learning framework.
We show that this simple method can greatly benefit downstream predictive tasks with reduced annotation cost.
Since the release of our work, similar image-text pretraining strategies have been used to improve more downstream healthcare tasks including image regeneration \cite{wang2021self}, medical visual question answering \cite{eslami2021does} and clinical risk prediction \cite{zang2021scehr}, etc.
Moreover, a similar idea can be extended to include other modalities of healthcare data, including multiomics data \cite{han2021pneumonia} or patient metadata \cite{vu2021medaug}, for more robust and cost-effective machine learning applications in the healthcare domain.

\section{Related Work}
\label{sec:related_work}

Our work is most relevant to work on medical image classification, which we have discussed in \refsec{intro}, and textual report generation from medical images \cite{wang2018tienet, jing2018automatic, liu2019clinically, miura2020improving}.
A dominant approach for initializing medical image encoders in relevant studies has been using encoder weights pretrained on ImageNet, despite the drastic difference in image characteristics \cite{raghu2019transfusion}.
Instead, we propose an alternative in-domain pretraining strategy for medical imaging and compare different pretraining approaches that also use the paired medical reports.
Our work is inspired by the recent line of work on image view-based contrastive learning \cite{henaff2020cpc,chen2020simple, he2020momentum, grill2020bootstrap,sowrirajan2021moco,azizi2021big}, but fundamentally differs from existing studies by exploiting contrastive learning using the text modality. As we show in \refsec{analysis}, the added semantics from the text data makes contrastive learning more effective in learning high-quality representations of medical images.
To our knowledge, our work represents the first systematic attempt in this direction.

% Recent work on image view-based contrastive visual representation learning has obtained promising results on general image classification and object recognition tasks \citep{chen2020simple, he2020momentum, grill2020bootstrap}.
% Our work is inspired by this line of work, 

Another line of work related to ours is visual-linguistic representation learning \cite{lu2019vilbert, tan2019lxmert, su2019vlbert}.
Among existing studies, \citet{ilharco2020probing} and \citet{gupta2020phrase} explored cross-modality contrastive objectives related to ours, but for the purpose of probing visual-linguistic models and learning phrase grounding, respectively.
Our work differs from most work in visual-linguistic pretraining in several crucial ways:
1) existing work in visual-linguistic learning focused on learning visual representations from paired text via a binary contrastive prediction task, whereas we contribute by showing the superior performance of the new cross-modality NCE objectives in improving visual representations;
2) existing work has primarily relied on object representations extracted from image segmentation models in their preprocessing steps, making them less applicable to medical image understanding tasks where anatomical segmentations are extremely hard to obtain;
3) while existing work has run evaluation primarily on visual-linguistic tasks such as visual question answering, we instead focus on evaluation with classification and retrieval tasks which are at the center of medical image understanding research.

Several concurrent papers have studied the problem of learning visual representations from text data \cite{sariyildiz2020learning, desai2021virtex} on general-domain image problems. Most notably, since the original release of our work, ConVIRT has been applied at larger scales in several general visual recognition studies, including the CLIP model \cite{radford2021learning}, which uses a simplified version of the ConVIRT approach,
and the ALIGN model by \citet{jia2021scaling}. These successful applications have confirmed that ConVIRT is a promising strategy for learning visual representations from human-written descriptive text, and that it has the potential to further advance the state of the art for visual recognition tasks. 

There are also subsequent studies which mainly focused on medical-domain image problems.
To the best of our knowledge, ConVIRT was the first work that leverages text-image contrastive loss for pretraining medical visual representations and was followed by numerous papers \cite{heiliger2022beyond} that apply multimodal contrastive learning to the medical imaging domain.
\citet{wang2021self} demonstrated the feasibility of such a pretraining strategy across mixed data inputs (image-only, text-only, image-text pairs) in three chest X-ray applications (i.e., classification, retrieval, and image regeneration).
\citet{muller2021joint} proposed a similar method, LoVT, for localized medical imaging tasks.
\citet{huang2021gloria} adapted our method and further proposed GloRIA to contrast image sub-regions and words in the paired report. 
\citet{liao2021multimodal} trained image and text encoders by encouraging the resulting representations to exhibit high local mutual information.
\citet{eslami2021does} proposed PubMedCLIP to better adapt CLIP to the Medical Visual Question Answering (MedVQA) task. 
\citet{zang2021scehr} applied a similar contrastive learning framework to clinical risk prediction based on longitudinal electronic health records. 
\citet{han2021pneumonia} extended ConVIRT to use radiomics features and contrastive learning for pneumonia detection, and \citet{vu2021medaug} selected positive pairs coming from views of possibly different images through the use of patient metadata.

\section{Methods}
\label{sec:method}

\subsection{Task Definition}

We start by defining our representation learning setting.
We assume paired input $(\bx_v, \bx_u)$ where $\bx_v$ represents one or a group of images, and $\bx_u$ represents a text sequence which describes the imaging information in $\bx_v$.
Our goal is to learn a parameterized image encoder function $f_v$, which maps an image to a fixed-dimensional vector.
We are then interested in transferring the learned image encoder function $f_v$ into downstream tasks, such as classification or image retrieval.
In this work, we model the encoder function $f_v$ as a convolutional neural network (CNN).

We note that paired image-text data $(\bx_v, \bx_u)$ naturally exists for many medical domains. Medical experts such as radiologists produce textual descriptions of images as part of their routine workflow, some of which are also made publicly available \cite{demner2016preparing, johnson2019mimic}.

\begin{figure*}
% \shrinkcaptionmargin
\centering
\includegraphics[width=0.95\textwidth]{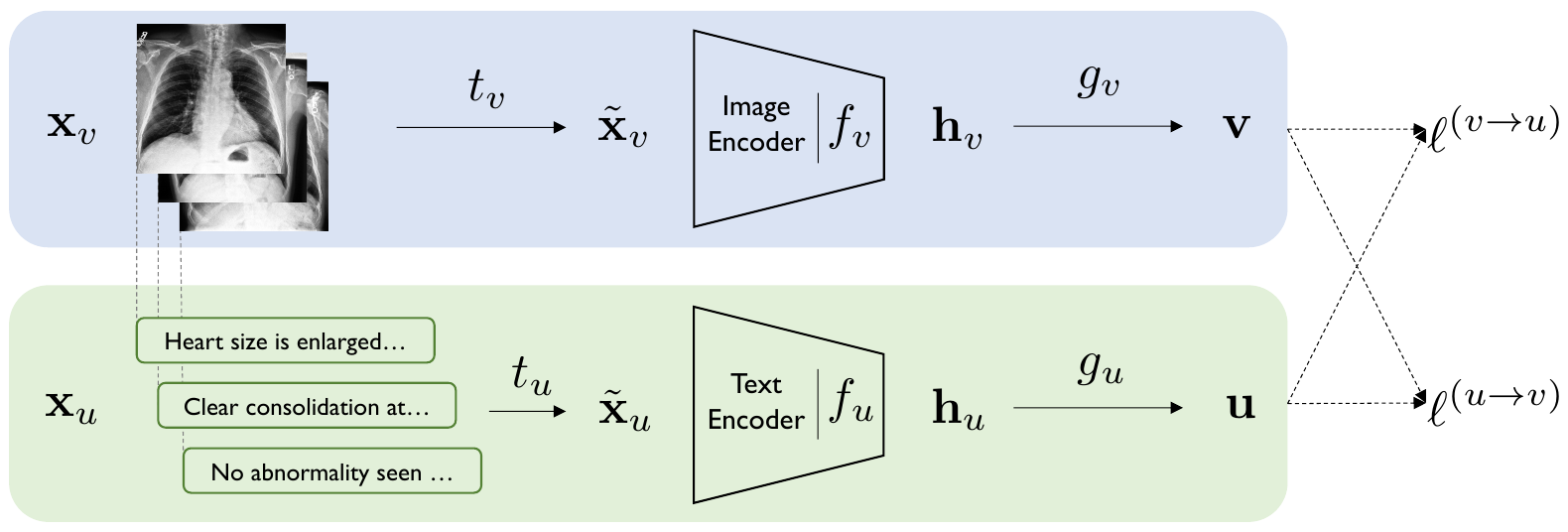}
\caption{
Overview of our ConVIRT framework.
The blue and green shades represent the image and text encoding pipelines, respectively.
% $t_v$ and $t_u$ represent transformation functions, and $g_v$ and $g_u$ represent projection functions for the two modalities.
Our method relies on maximizing the agreement between the true image-text representation pairs with bidirectional losses $\ell^{(v \rightarrow u)}$ and $\ell^{(u \rightarrow v)}$.
}
% \halfshrinkcaptionmargin
\label{fig:diagram}
\end{figure*}

\subsection{Contrastive Visual Representation Learning from Text}

An overview of our method, ConVIRT, for learning $f_v$ is shown in \reffig{diagram}.
At a high level, our method converts each input image $\bx_v$ and text $\bx_u$ into $d$-dimensional vector representations $\bv$ and $\bu$ respectively, following a similar processing pipeline.
For each input image $\bx_v$, our method starts by drawing a random view $\tilde{\bx}_v$ from $\bx_v$ with a sampled transformation function $t_v \sim \sT$, where $\sT$ represents a family of stochastic image transformation functions described later.
Next, the encoder function $f_v$ transforms $\tilde{\bx}_v$ into a fixed-dimensional vector $\bh_v$, followed by a non-linear projection $g_v$ which further transforms $\bh_v$ into vector $\bv$:
\begin{equation}
    \bv = g_v ( f_v (\tilde{\bx}_v)),
\end{equation}
where $\bv \in \BR^d$.
Similarly, for each text input $\bx_u$, we obtain a span $\tilde{\bx}_u$ from it following a sampling function $t_u$, and then a text representation $\bu$ with:
$\bu = g_u ( f_u (\tilde{\bx}_u))$,
where $f_u$ is a text encoder, $g_u$ a projection, and $\bu \in \BR^d$.
The projection functions $g_v$ and $g_u$ project representations for both modalities from their encoder space to the same $d$-dimensional space for contrastive learning.

At training time, we sample a minibatch of $N$ input pairs ($\bx_v$, $\bx_u$) from training data, and calculate their representation pairs ($\bv$, $\bu$).
We use ($\bv_i$, $\bu_i$) to denote the $i$-th pair.
The training objective of ConVIRT involves two loss functions.
The first loss function is an image-to-text contrastive loss for the $i$-th pair:
\begin{equation}
    \ell^{(v \rightarrow u)}_{i} = - \log \frac{ \exp(\langle \bv_i, \bu_i \rangle / \tau) }{ \sum^{N}_{k=1} \exp (\langle \bv_i, \bu_k \rangle / \tau)},
\end{equation}
where $\langle \bv_i, \bu_i \rangle$ represents the cosine similarity, i.e., $\langle \bv, \bu \rangle = \bv^\top \bu / \| \bv \| \| \bu \|$; and $\tau \in \R^+$ represents a temperature parameter.
This loss takes the same form as the InfoNCE loss \cite{oord2018representation}, and minimizing it leads to encoders that maximally preserve the mutual information between the true pairs under the representation functions.
Intuitively, it is the log loss of an $N$-way classifier that tries to predict ($\bv_i$, $\bu_i$) as the true pair.
Note that unlike previous work which use a contrastive loss between inputs of the same modality \cite{chen2020simple, he2020momentum}, our image-to-text contrastive loss is asymmetric for each input modality.
We therefore define a similar text-to-image contrastive loss as:
\begin{equation}
    \ell^{(u \rightarrow v)}_{i} = - \log \frac{ \exp(\langle \bu_i, \bv_i \rangle / \tau) }{ \sum^{N}_{k=1} \exp (\langle \bu_i, \bv_k \rangle / \tau)}.
\end{equation}
Our final training loss is then computed as a weighted combination of the two losses averaged over all positive image-text pairs in each minibatch:
\begin{equation}
    \sL = \frac{1}{N} \sum^{N}_{i=1} \Big( \lambda \ell^{(v \rightarrow u)}_{i} + (1-\lambda) \ell^{(u \rightarrow v)}_{i} \Big),
\end{equation}
where $\lambda \in [0,1]$ is a scalar weight.

\subsection{Realization}

We note that our ConVIRT framework defined above is agnostic to the specific choice of image and text encoders, transformations and projection functions.
Following previous work \cite{chen2020simple}, we model $g_v$ and $g_u$ as separate learnable single-hidden-layer neural networks, i.e., $g_v (\cdot) = \bW^{(2)} \sigma (\bW^{(1)} (\cdot))$ where $\sigma$ is a ReLU non-linearity, and similarly for $g_u$.

For the image encoder $f_v$, we use the ResNet50 architecture \cite{he2016deep} for all experiments, as it is the architecture of choice for much medical imaging work and is shown to achieve competitive performance.
For the text encoder $f_u$, we use a BERT encoder \cite{devlin2019bert} followed by a max-pooling layer over all output vectors.
% \footnote{We also experimented with using a mean-pooling layer or using the special [CLS] token representation from BERT as our pooling strategy \citep{reimers2019sentence}, and found max-pooling achieved the best overall performance, and therefore used it consistently across all experiments.}
We initialize our encoder with the ClinicalBERT weights \cite{alsentzer2019publicly} pretrained on the MIMIC clinical notes, which achieved state-of-the-art performance on a suite of clinical NLP tasks.
At training time we allow the encoder to adapt to our contrastive task by freezing the embeddings and the first 6 transformer layers of this BERT encoder and fine-tuning the last 6 layers.

For the image transformation family $\sT$ where $t_v$ is sampled from, we use sequential applications of five random transformations: \emph{cropping}, \emph{horizontal flipping}, \emph{affine transformation}, \emph{color jittering} and \emph{Gaussian blur}.
% We adopted the torchvision%
% \footnote{\url{https://github.com/pytorch/vision}}
% implementations for all transformations, and 
Different from recent work on contrastive visual learning \cite{chen2020simple, chen2020improved}, we only apply brightness and contrast adjustments in \emph{color jittering}, due to the monochrome nature of the medical images.
For the text transformation function $t_u$, we apply a simple uniform sampling of a sentence from the input document $\bx_u$ (i.e., $\tilde{\bx}_u$ is a randomly sampled sentence from $\bx_u$ for each minibatch).
We did not use a more aggressive transformation mainly because sampling at the sentence level helps preserve the semantic meaning of the sampled spans.

An alternative method to using the sampled view $\tilde{\bx}_v$ from $\bx_v$ as input to the encoder is to directly use $\bx_v$ or to fuse all images for each study in the case of multiple available $\bx_v$ instances (e.g., images from multiple angles).
We empirically found in our preliminary experiments that using sampled view $\tilde{\bx}_v$ leads to better pretraining results.
We conjecture that we can treat the use of $\tilde{\bx}_v$ as a way of data augmentation for the visual modality, which helped increase the effective amount of unique image-text pairs that the model sees at pretraining time, leading to better performance.

\section{Experiments}
\label{sec:experiments}

We now introduce the paired datasets that we used for contrastive pretraining, the downstream tasks and datasets for evaluation, and the baseline methods that we compare against.
% We introduce our pretraining and evaluation datasets and the baselines that we compare against.

\subsection{Data for Pretraining}

We evaluate ConVIRT by pretraining two separate image encoders using two separate image-text datasets (see \refapp{pretraining-details} for full pretraining details):

\begin{itemize}[leftmargin=*,align=left]
\item \textbf{Chest} image encoder: We use version 2 of the public \textbf{MIMIC-CXR} database \cite{johnson2019mimic}, which is a collection of chest radiograph images paired with their textual reports, and since its release has become a standard resource for studying multi-modal modeling of medical images.
After preprocessing, this dataset contains a total of about 217k image-text pairs, with each pair containing an average of 1.7 images and 6.0 sentences.
% We use this dataset for the pretraining of our chest medical imaging encoders.

\item \textbf{Bone} image encoder: We obtain a collection of musculoskeletal (i.e., bone) image-text pairs from the Rhode Island Hospital system.
Following chest, musculoskeletal images constitute the second most common type of radiograph images in a typical hospital.
This dataset contains a total of 48k image-text pairs, with each pair containing an average of 2.5 images and 8.0 sentences.
% We use this dataset for the pretraining of our bony medical imaging encoders.
\end{itemize}

% We include pretraining details in \refapp{pretraining-details}.

\subsection{Evaluation Tasks \& Data}

We evaluate our pretrained image encoders on three medical imaging tasks: image classification, zero-shot image-image retrieval and zero-shot text-image retrieval.
% We now describe each of the evaluation settings as well as the datasets used.

\paragraph{Image Classification.}
We evaluate our pretrained image encoders on four representative medical image classification tasks:
1) \textbf{RSNA Pneumonia Detection} \cite{wang2017chestx,shih2019augmenting}, which involves binary classification of a chest radiograph image into either a \emph{pneumonia} or a \emph{normal} category;
2) \textbf{CheXpert} image classification \cite{irvin2019chexpert}, which involves multi-label binary classification of a chest image for five individual labels, i.e., \emph{atelectasis}, \emph{cardiomegaly}, \emph{consolidation}, \emph{edema} and \emph{pleural effusion};
3) \textbf{COVIDx} \cite{wang2020covid}, which involves multi-class chest image classification into three categories (\emph{COVID19}, \emph{non-COVID pneumonia} or \emph{normal});
and 4) \textbf{MURA} bony abnormality detection \cite{rajpurkar2017mura}, which involves binary classification of a musculoskeletal image into \emph{abnormal} or \emph{normal}.
We report test accuracy for COVIDx given its balanced test set, and report the standard area under the receiver operating characteristic curve (AUC) metric for other tasks.

Following previous work \cite{henaff2020cpc, chen2020simple, he2020momentum}, for all tasks, we evaluate each pretrained image encoder under two settings: a \textbf{linear classification} setting, where the pretrained CNN weights are frozen and only a linear classification head is trained for the task; and a \textbf{fine-tuning} setting, where both the CNN weights and the linear head are fine-tuned.
The two settings complement each other for evaluation purposes: while the linear setting directly evaluates the quality of the extracted image features with the pretrained CNN, the fine-tuning setting more closely resembles how the pretrained CNN weights are used in practical applications.

To further compare the \textbf{data efficiency} of different pretraining methods, for each setting we evaluate the image encoders with \textbf{1\%}, \textbf{10\%} and \textbf{all} training data, respectively (except for the COVIDx task where we omit the 1\% setting due to the scarcity of training data).
To control the variance in results, for all settings and models, we report \textbf{average results} over 5 independent training runs.
We include further dataset and training details in \refapp{classification-details}.

\paragraph{Zero-shot Image-image Retrieval.}
This evaluation is similar to the conventional content-based image retrieval setting in which we search for images of a particular category using a representative \emph{query} image.
For evaluation, a group of query images and a larger collection of \emph{candidate} images, each with a categorical label, are given to a pretrained CNN encoder.
We encode each query and candidate image with this encoder, and then for each query, rank all candidates by their cosine similarities to the query in descending order.
Since a widely-used annotated benchmark for this setting is not available, we create our own dataset by re-using existing annotations in the CheXpert dataset \cite{irvin2019chexpert} and additional expert annotations from a board-certified radiologist.
The resulting dataset covers 8 different chest abnormality categories, each with 10 expert-annotated query and 200 candidate images.
We include the detailed collection and annotation procedure in \refapp{image-image}, and refer to this dataset as \textbf{CheXpert $\mathbf{8{\times}200}$ Retrieval Dataset}.
We focus our evaluation on retrieval precision, and evaluate our models with Precision@$k$ metrics where $k=5, 10, 100$.

\paragraph{Zero-shot Text-image Retrieval.}
This setting is similar to the image-image retrieval setting, but instead of using query images, we retrieve images of a particular category with textual queries.
For this purpose, we ask a radiologist to write 5 diverse and representative textual descriptions for each of the 8 abnormality categories for the same CheXpert 8x200 candidate images (see \refapp{text-image} for details).
At test time, for each query we encode its text with the learned text encoder $f_u$ and then retrieve from candidate images in a similar way.
This evaluation not only evaluates the quality of the learned image representations, but also the alignment between the text representations and the image representations.
We again use Precision@$k$ metrics where $k=5, 10, 100$.

% We note that both retrieval evaluation settings are \textbf{zero-shot} in nature, because we directly use the pretrained CNN weights for evaluation and never fine-tune the weights for the particular tasks.

\subsection{Baseline Methods}
We compare ConVIRT against the following standard or competitive initialization methods:
\begin{itemize}[leftmargin=*,align=left]
\setlength\itemsep{0em}
\item \textbf{Random Init.}:
For all tasks we initialize the ResNet50 with its default random initialization.

\item \textbf{ImageNet Init.}:
We use CNN weights pretrained on ImageNet \cite{russakovsky2015imagenet}, which remains a dominant initialization approach for medical imaging work \cite{raghu2019transfusion}.
% We initialize the ResNet50 with the weights pretrained on the standard ImageNet ILSVRC-2012 task \citep{russakovsky2015imagenet}.
% We include this as a baseline since ImageNet pretraining remains a dominant approach for medical imaging work \citep{raghu2019transfusion}.

\item \textbf{Caption-LSTM}:
We further pretrain the ImageNet-initialized CNN weights with an image captioning task using the standard CNN-LSTM with attention model \cite{xu2015show}.
We train the model to decode the paired medical report text from the encoded image representations.
Compared to the random or ImageNet initializations, this is an ``in-domain'' initialization baseline which uses the paired text data for representation learning.

\item \textbf{Caption-Transformer}:
We use a CNN-Transformer-based captioning model \cite{cornia2020meshed} for caption-based pretraining, which recently achieves state-of-the-art results on the COCO image captioning benchmark \cite{lin2014microsoft}.

\item \textbf{Contrastive-Binary-Loss}:
This baseline differs from ConVIRT by contrasting the paired image and text representations with a binary classification head, as is widely done in visual-linguistic pretraining work \cite{tan2019lxmert, su2019vlbert}.
For each input pair, we first project encoder outputs $\bh_v$ and $\bh_u$ into the same dimension with linear layers, concatenate them, and use a MLP network to predict a binary probability of whether the input is a real or a ``fake'' pair, which we train with a binary cross-entropy loss.
During training, for each ($\bx_v$, $\bx_u$) pair in the training set, we construct a ``fake'' pair by replacing $\bx_u$ with a randomly sampled one from the dataset.
We expect that the binary classification task requires the encoder to learn reasonable representations of the input images, and therefore is a stronger in-domain initialization baseline.

\end{itemize}

For fair comparison, for all baselines that require paired image-text data, we use the same datasets as in our contrastive pretraining.
For the captioning-based methods, we always use the model checkpoints that achieve the best CIDEr score \cite{vedantam2015cider} on a held-out validation set.

\section{Results} 
\label{sec:results}

\subsection{Classification Tasks}

\paragraph{Linear Classification.}
We present all linear classification results for the four tasks in \reftab{classification}(a).
We find that compared to random initialization, ImageNet initialization provides markedly better representations, despite pretrained on a very different domain of images;
in-domain image initialization methods that use paired image-text data further improve over ImageNet initialization in almost all settings.
Among the in-domain initialization methods, our proposed ConVIRT pretraining achieves the best overall results in all settings.
Notably, we find on three out of the four tasks, with only 1\% training data ConVIRT is able to achieve classification results better than the default ImageNet initialization with 100\% training data, highlighting the high quality of the learned representations from ConVIRT.

\paragraph{Fine-tuning.}
We show the fine-tuning evaluation results in \reftab{classification}(b).
Similar to the linear setting, we find that:
1) ImageNet initialization is again better than random initialization with smaller margins;
2) all in-domain initialization methods are better than the popular ImageNet initialization in most settings;
and 3) our proposed ConVIRT pretraining again achieves the best overall results in 10 out of the 11 settings, with the exception of the CheXpert dataset with all training data used, where the result of ConVIRT is similar to that of the Caption-Transformer result.
Most notably, on all datasets, with only 10\% labeled training data ConVIRT achieves classification results that are better or close to the ImageNet initialization with 100\% training data results.

We also notice that our conclusion of using ImageNet versus random initialization is different from \cite{raghu2019transfusion}:
while they showed comparable results from the two strategies, we find that using ImageNet initialization is still superior than random initialization in most results, justifying its popularity.
Upon closer examination, we conjecture that this is likely due to under-optimization of their models: while our ResNet50 with random initialization achieves an average AUC of 85.8 on the CheXpert dataset, their ResNet50 model only achieved 83.5 AUC on the same evaluation set.

\begin{table}[t]
% \begin{subtable}[t]{2in}
\begin{center}
\small
\setlength{\tabcolsep}{0.8em}

\subtable[Linear classification]{
\label{tab:linear}
% \begin{tabular}{l@{\hskip 2em}ccc@{\hskip 2em}ccc@{\hskip 2em}cc@{\hskip 2em}ccc}
\resizebox{\textwidth}{!}{
\begin{tabular}{lccccccccccc}
\toprule
& \multicolumn{3}{c}{RSNA (AUC)} & \multicolumn{3}{c}{CheXpert (AUC)} & \multicolumn{2}{c}{COVIDx (Accu.)} & \multicolumn{3}{c}{MURA (AUC)} \\
Method  & 1\% & 10\% & all & 1\% & 10\% & all & 10\% & all & 1\% & 10\% & all\\
\midrule
\multicolumn{12}{l}{\textit{General initialization methods}}\\
Random Init. & 55.0 & 67.3 & 72.3 & 58.2 & 63.7 & 66.2 & 69.2 & 73.5 & 50.9 & 56.8 & 62.0 \\
ImageNet Init. & 82.8 & 85.4 & 86.9 & 75.7 & 79.7 & 81.0 & 83.7 & 88.6 & 63.8 & 74.1 & 79.0 \\
\midrule
\multicolumn{12}{l}{\textit{In-domain initialization methods}}\\
Caption-Transformer & 84.8 & 87.5 & 89.5 & 77.2 & 82.6 & 83.9 & 80.0 & 89.0 & 66.5 & 76.3 & 81.8 \\
Caption-LSTM & 89.8 & 90.8 & 91.3 & 85.2 & 85.3 & 86.2 & 84.5 & \bf{91.7} & 75.2 & 81.5 & 84.1 \\
Contrastive-Binary-Loss & 88.9 & 90.5 & 90.8 & 84.5 & 85.6 & 85.8 & 80.5 & 90.8 & 76.8 & 81.7 & 85.3 \\
ConVIRT (Ours) & \cc \bf{90.7} & \cc \bf{91.7} & \cc \bf{92.1} & \cc \bf{85.9} & \cc \bf{86.8} & \cc \bf{87.3} & \cc \bf{85.9} & \cc \bf{91.7} & \cc \bf{81.2} & \cc \bf{85.1} & \cc \bf{87.6} \\
\bottomrule
\end{tabular}
}
% \shrinkcaptionmargin
% \caption{Linear classification}
}
\end{center}

% \end{subtable}

% \vspace{0.5em}
% \begin{subtable}[t]{2in}
\begin{center}
\small
\setlength{\tabcolsep}{0.8em}

\subtable[Fine-tuning]{
\label{tab:finetune}
\resizebox{\textwidth}{!}{
\begin{tabular}{lccccccccccc}
\toprule
& \multicolumn{3}{c}{RSNA (AUC)} & \multicolumn{3}{c}{CheXpert (AUC)} & \multicolumn{2}{c}{COVIDx (Accu.)} & \multicolumn{3}{c}{MURA (AUC)} \\
Method  & 1\% & 10\% & all & 1\% & 10\% & all & 10\% & all & 1\% & 10\% & all\\
\midrule
\multicolumn{12}{l}{\textit{General initialization methods}}\\
Random Init. & 71.9 & 82.2 & 88.5 & 70.4 & 81.1 & 85.8 & 75.4 & 87.7 & 56.8 & 61.6 & 79.1 \\
ImageNet Init. & 83.1 & 87.3 & 90.8 & 80.1 & 84.8 & 87.6 & 84.4 & 90.3 & 72.1 & 81.8 & 87.0 \\
\midrule
\multicolumn{12}{l}{\textit{In-domain initialization methods}}\\
Caption-Transformer & 86.3 & 89.2 & 92.1 & 81.5 & 86.4 & \bf{88.2} & 88.3 & 92.3 & 75.2 & 83.2 & 87.6 \\
Caption-LSTM & 87.2 & 88.0 & 91.0 & 83.5 & 85.8 & 87.8 & 83.8 & 90.8 & 78.7 & 83.3 & 87.8 \\
Contrastive-Binary-Loss & 87.7 & 89.9 & 91.2 & 86.2 & 86.1 & 87.7 & 89.5 & 90.5 & 80.6 & 84.0 & 88.4 \\
ConVIRT (Ours) & \cc \bf{88.8} & \cc \bf{91.5} & \cc \bf{92.7} & \cc \bf{87.0} & \cc \bf{88.1} & \cc 88.1 & \cc \bf{90.3} & \cc \bf{92.4} & \cc \bf{81.3} & \cc \bf{86.5} & \cc \bf{89.0} \\
\bottomrule
\end{tabular}
}
% \shrinkcaptionmargin
% \caption{Fine-tuning}
}
\end{center}

% \halfshrinkcaptionmargin
\caption{Results for the medical image classification tasks: (a) linear classification; (b) fine-tuning setting.
All results are averaged over 5 independent models.
Best results for each setting are in boldface.
% Results higher than the second best by $\geq 0.5$ are in boldface.
COVIDx 1\% setting is omitted due to the scarcity of labels in COVIDx.
}
% \halfshrinkcaptionmargin
\label{tab:classification}
\end{table}

\begin{table*}[t]
\begin{center}
\small
\setlength{\tabcolsep}{0.8em}

\resizebox{\textwidth}{!}{
\begin{tabular}{lcccccc}
\toprule
& \multicolumn{3}{c}{Image-Image Retrieval} & \multicolumn{3}{c}{Text-Image Retrieval} \\
Method  & Prec@5 & Prec@10 & Prec@50 & Prec@5 & Prec@10 & Prec@50 \\
\midrule
% \multicolumn{4}{l}{\textit{General initialization methods}}\\
Random & 12.5 & 12.5 & 12.5 & 12.5 & 12.5 & 12.5 \\
ImageNet & 14.8 & 14.4 & 15.0 & -- & -- & --\\
\midrule
\multicolumn{4}{l}{\textit{In-domain initialization methods}}\\
Caption-Transformer & 29.8 & 28.0 & 23.0 & -- & -- & -- \\
Caption-LSTM & 34.8 & 32.9 & 28.1 & -- & -- & --\\
Contrastive-Binary-Loss & 38.8 & 36.6 & 29.7 & 15.5 & 14.5 & 13.7\\
ConVIRT (Ours) & \cc \bf{45.0} & \cc \bf{42.9} & \cc \bf{35.7} & \cc \bf{60.0} & \cc \bf{57.5} & \cc \bf{48.8} \\
\midrule
\multicolumn{4}{l}{\textit{Fine-tuned}}\\
ConVIRT + CheXpert Supervised & 56.8 & 56.3 & 48.9 & -- & -- & --
\\
\bottomrule
\end{tabular}
}
\end{center}
% \shrinkcaptionmargin
\caption{Zero-shot image-image and text-image retrieval results on the CheXpert $8{\times}200$ datasets.
\emph{Random} shows results from a random guess; \emph{ConVIRT + CheXpert Supervised} shows results from further fine-tuning the pretrained weights with supervised training data.
Text-image retrieval results are not obtained for some methods due to the lack of text encoders.
}
% \shrinkcaptionmargin
\label{tab:retrieval}
\end{table*}

\subsection{Retrieval Tasks}

We present the zero-shot image-image and text-image retrieval results in \reftab{retrieval}.
For the image-image retrieval setting, we present additional results from fine-tuning our pretrained model on all CheXpert training data, and use them as ``upper bounds'' of the results obtained from the use of supervised labels.
We find that:
1) using ImageNet weights in a zero-shot image retrieval setting is only better than random guess by small margins;
2) all in-domain pretrained CNN weights achieve much better retrieval performance than ImageNet weights;
and 3) our proposed ConVIRT pretraining achieves the best overall retrieval results on all metrics.
While Contrastive-Binary-Loss performs notably better than other baselines in image-image retrieval, its text-image retrieval results are far from ConVIRT pretraining.
We conjecture that the lack of an explicit similarity-based loss function in the Contrastive-Binary-Loss baseline results in misaligned representations in the image and text space, leading to poor results in text-image retrieval.

% We additionally run t-SNE visualizations \citep{maaten2008visualizing} of 1000 candidate images from 5 different categories in the retrieval dataset, and find that ConVIRT leads to more clearly separated images in the encoding space. We include the visualization results in \refapp{tsne}.

\begin{figure}
\small
% \hfill
\centering
\subfigure[ImageNet Pretraining][t] {
\centering
\includegraphics[width=0.3\linewidth]{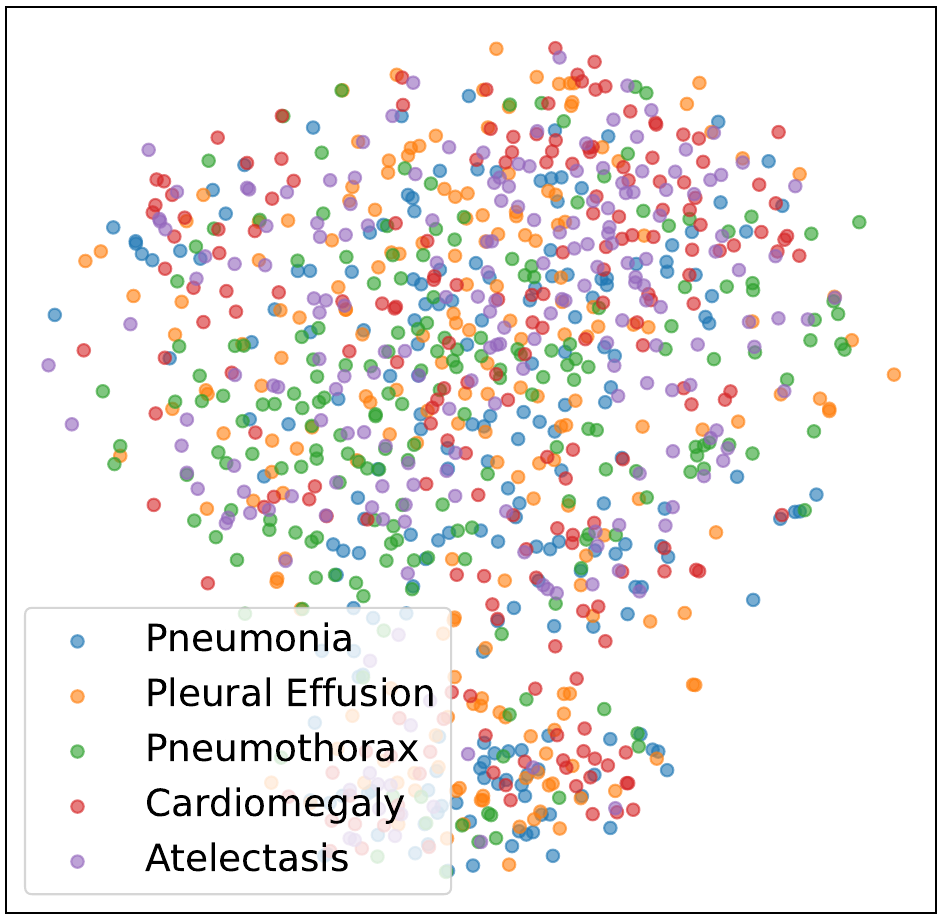}
% \caption{ImageNet Pretraining}
}\qquad
\subfigure[ConVIRT Pretraining][t]{
\centering
\includegraphics[width=0.3\linewidth]{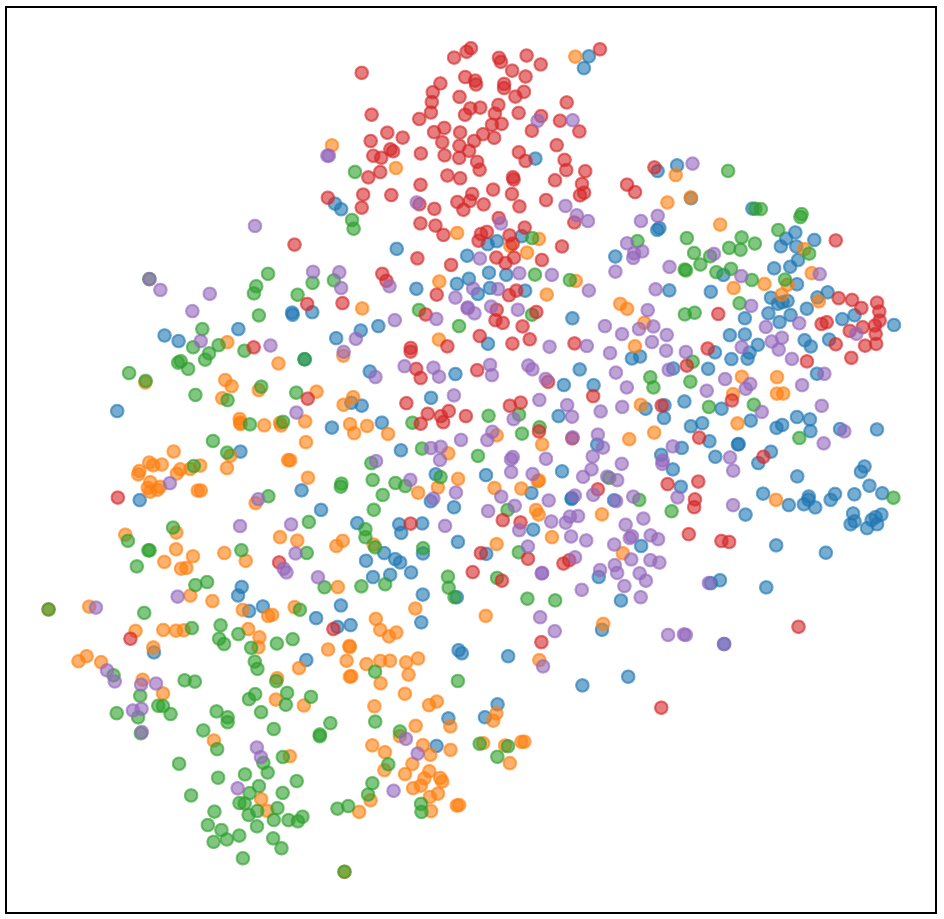}
}
% \hfill
% \halfshrinkcaptionmargin
\caption{t-SNE visualizations of encoded image representations from different pretraining methods.}
% \shrinkcaptionmargin
\label{fig:tsne}
\end{figure}

To understand how well ConVIRT pretraining helps separate images from different abnormality categories in its encoding space, in \reffig{tsne} we present t-SNE plots \citep{maaten2008visualizing} of candidate images in the CheXpert 8x200 dataset for five selected categories, from the ImageNet pretrained CNN encoder and the ConVIRT pretrained encoder.
It is worth noting that clustering images in our setting is much more challenging than that in the general object classification setting due to the high inter-class similarity of the medical images.
Nevertheless we find that ConVIRT pretraining achieves a better clustering of the images in the t-SNE plots.

% t-SNE plots

\section{Analysis and Discussion}
\label{sec:analysis}

% In this section we present analysis and discussions about factors that influence the performance of our proposed method and its comparisons to existing image-only unsupervised pretraining methods.

\paragraph{Comparisons to Image-only Contrastive Learning.}
ConVIRT shows superior results against baselines in evaluation, but an important question remains as to how it compares against existing image-only contrastive learning methods.
We study this by running two popular such methods, SimCLR \cite{chen2020simple} and MoCo v2 \cite{chen2020improved}, on the same collection of images that we used in our pretraining.
We present the results in \reftab{unsupervised} and include training details in \refapp{image-only}.
We find that compared to ImageNet initialization, both contrastive methods lead to marginal to moderate improvements on the classification and retrieval tasks.
However, our training strategy substantially outperforms both methods on all tasks, demonstrating its effective use of information from the paired text data. This efficient use of data is critical to the healthcare domain because medical data are often limited in size but come with paired text data and even user metadata.

% comparison to image-only learning
\begin{table}[t]
\begin{center}
% \small
\setlength{\tabcolsep}{0.2em}
\begin{tabular}{lccccc}
\toprule
& RSNA & CheXpert & Image-Image \\
Method  & (Linear, 1\%) & (Linear, 1\%) & (Prec@10) \\
\midrule
ImageNet & 82.8 & 75.7 & 14.4 \\
\midrule
SimCLR \cite{chen2020simple} & 86.3 & 77.4 & 17.6 \\
MoCo v2 \cite{chen2020improved} & 86.6 & 81.3 & 20.6 \\
\midrule
ConVIRT & 90.7 & 85.9 & 42.9 \\
\bottomrule
\end{tabular}
\end{center}
% \shrinkcaptionmargin
\caption{Comparisons of ConVIRT to image-only contrastive learning.
For RSNA and CheXpert we show the AUC under linear classification with 1\% training data.}
% \shrinkcaptionmargin
\label{tab:unsupervised}
\end{table}

% saliency maps
\begin{figure*}[t]
\centering
\includegraphics[width=0.95\textwidth]{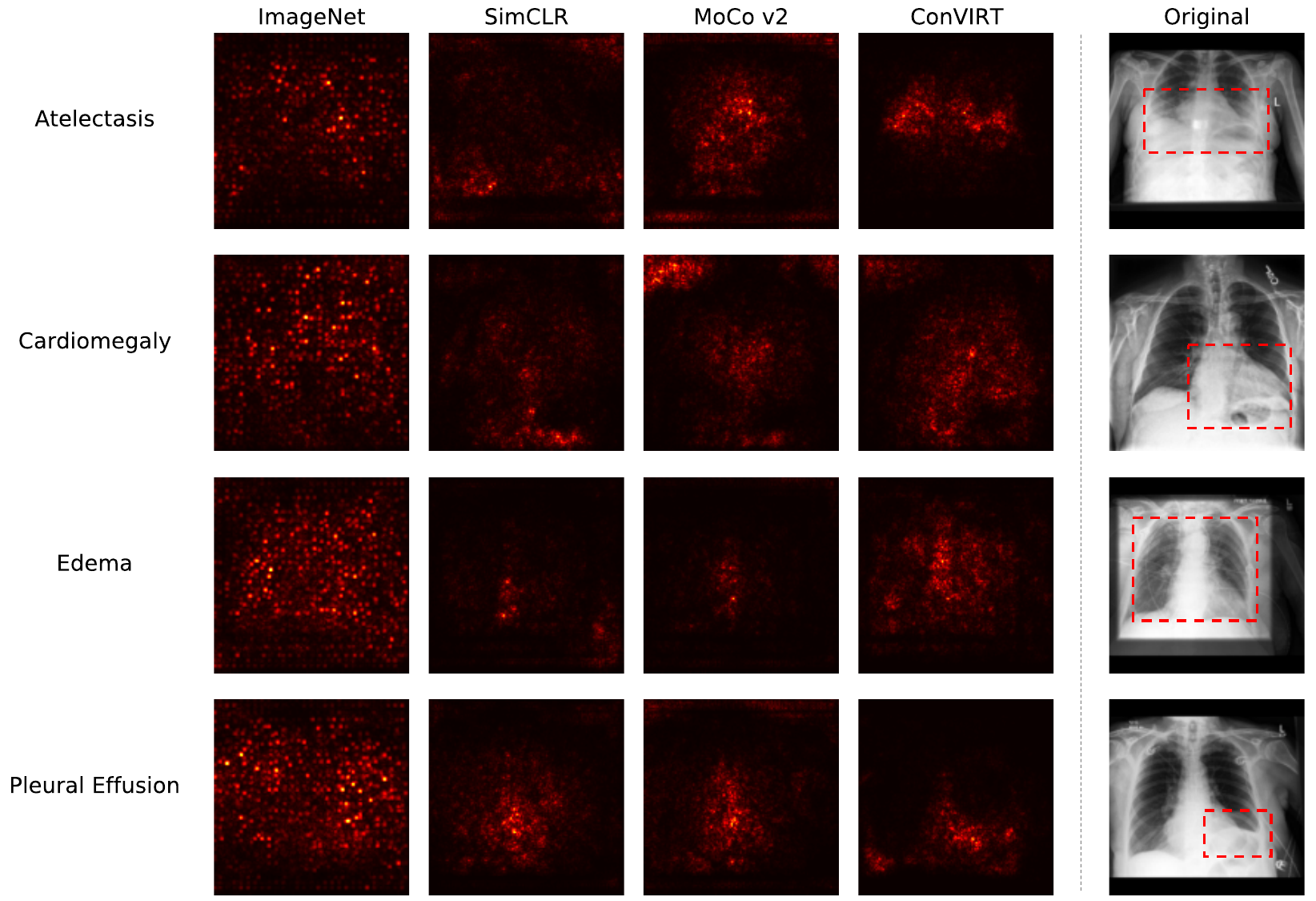}
% \shrinkcaptionmargin
\caption{
Saliency maps on sampled images for 4 abnormality categories in the CheXpert dataset.
For each image we present maps for ImageNet, SimCLR, MoCo v2 and our ConVIRT initializations.
Ground truth regions that are indicative of the abnormalities are shown as red boxes in the original images on the right, and are seen to most closely match the regions found by ConVIRT.
}
% \shrinkcaptionmargin
\label{fig:saliency}
\end{figure*}

To understand the representational difference that has led to this difference in performance, for all four initialization methods, we visualize in \reffig{saliency} the saliency maps \cite{simonyan2014saliency} corresponding to the correct class on sampled images from the CheXpert dataset.
Models for all initialization methods are trained with 1\% CheXpert training data under the linear classification setting (with pretrained CNN weights frozen).
We find that ImageNet pretraining has led to models that focus on trivial visual features that are mostly irrelevant to the task, and that the model with ConVIRT pretrained weights has focused on much more relevant areas than those with SimCLR and MoCo v2 pretraining, suggesting more effective representation learning.
For example, for \emph{atelectasis}, while the ConVIRT model has correctly focused on the bottom of the lung regions, the SimCLR model has much more scattered focus and the MoCo model has incorrectly focused on the heart region.

% pretraining correlation
\begin{figure*}
\centering
% \small
% \shrinkcaptionmargin
\pgfplotstableread[row sep=\\,col sep=&]{
epoch & loss & a & b & c \\
5 & -2.869 & 26.62 & 40.5 & 89.578 \\
10 & -2.759 & 31.38 & 41.75 & 89.64 \\
15 & -2.695 & 31.88 & 41.75 & 89.884 \\
20 & -2.677 & 34.88 & 45.75 & 89.716 \\
25 & -2.622 & 34.75 & 44 & 89.872 \\
30 & -2.586 & 35.25 & 52.5 & 89.65 \\
35 & -2.555 & 36.62 & 57.25 & 89.592 \\
40 & -2.571 & 39.12 & 47.25 & 90.114 \\
45 & -2.537 & 37.25 & 47.5 & 89.83 \\
50 & -2.512 & 39.00 & 47.75 & 89.942 \\
55 & -2.501 & 37.00 & 48.75 & 90.324 \\
60 & -2.479 & 37.62 & 51.75 & 90.238 \\
65 & -2.462 & 39.13 & 55.5 & 89.902 \\
70 & -2.450 & 34.62 & 52.5 & 90.108 \\
75 & -2.437 & 37.00 & 50 & 90.262 \\
80 & -2.394 & 38.38 & 57.25 & 90.218 \\
85 & -2.377 & 39.50 & 53.5 & 90.1 \\
90 & -2.367 & 39.62 & 48.5 & 90.492 \\
95 & -2.369 & 43.12 & 55.5 & 90.366 \\
100 & -2.355 & 39.88 & 50.75 & 90.532 \\
105 & -2.351 & 39.13 & 55 & 90.60 \\
110 & -2.330 & 39.25 & 58 & 90.68 \\
115 & -2.328 & 41.50 & 54 & 90.54 \\
120 & -2.317 & 41.00 & 56 & 90.38 \\
125 & -2.314 & 42.62 & 60.25 & 90.66 \\
130 & -2.311 & 42.50 & 56.25 & 90.71 \\
135 & -2.304 & 41.50 & 57.25 & 90.62 \\
140 & -2.306 & 42.12 & 58.5 & 90.60 \\
145 & -2.304 & 40.75 & 59.25 & 90.58 \\
150 & -2.299 & 43.25 & 59 & 90.55 \\
155 & -2.302 & 42.37 & 58.25 & 90.61 \\
160 & -2.297 & 42.62 & 59.5 & 90.66 \\
165 & -2.296 & 41.38 & 59.75 & 90.62 \\
170 & -2.294 & 42.62 & 58.5 & 90.63 \\
175 & -2.293 & 41.75 & 57.5 & 90.61 \\
180 & -2.294 & 43.12 & 57.25 & 90.66 \\
185 & -2.293 & 41.62 & 58.25 & 90.69 \\
190 & -2.292 & 42.75 & 56 & 90.69 \\
195 & -2.291 & 42.00 & 56.75 & 90.67 \\
200 & -2.291 & 42.12 & 58 & 90.68 \\
}\alldata

\pgfplotstableread[row sep=\\,col sep=&]{
epoch & loss \\
5 & 2.869 \\
10 & 2.759 \\
15 & 2.695 \\
20 & 2.677 \\
25 & 2.622 \\
30 & 2.586 \\
35 & 2.555 \\
40 & 2.571 \\
45 & 2.537 \\
50 & 2.512 \\
55 & 2.501 \\
60 & 2.479 \\
65 & 2.462 \\
70 & 2.450 \\
75 & 2.437 \\
80 & 2.394 \\
85 & 2.377 \\
90 & 2.367 \\
95 & 2.369 \\
100 & 2.355 \\
105 & 2.351 \\
110 & 2.330 \\
115 & 2.328 \\
120 & 2.317 \\
125 & 2.314 \\
130 & 2.311 \\
135 & 2.304 \\
140 & 2.306 \\
145 & 2.304 \\
150 & 2.299 \\
155 & 2.302 \\
160 & 2.297 \\
165 & 2.296 \\
170 & 2.294 \\
175 & 2.293 \\
180 & 2.294 \\
185 & 2.293 \\
190 & 2.292 \\
195 & 2.291 \\
200 & 2.291 \\
}\lossdata

\pgfplotsset{width=0.3\textwidth, compat=1.9, height=1.6in}
\newcommand{\tickfontsize}{\scriptsize}

\begin{tikzpicture}[every plot/.append style={thick},font=\small]
\matrix[inner sep=2mm]{
\begin{axis}[
xlabel style={align=left},
xlabel={(a) Pretraining Loss},
tick label style={font=\tickfontsize},
yticklabel style={xshift=1ex}
]
\addplot[color=red, mark=*, mark size=0.15em] table[x=epoch, y=loss]{\lossdata};
\end{axis}

&
\begin{axis}[
xlabel style={align=center},
xlabel={(b) RSNA Linear \\(1\%, AUC)},
ymin=89.3,
ymax=90.75,
tick label style={font=\tickfontsize},
yticklabel style={xshift=1ex}
]
\addplot[color=orange, mark=*, mark size=0.15em, only marks] table[x=loss, y=c]{\alldata};
\end{axis}

&
\begin{axis}[
xlabel style={align=center},
xlabel={(c) Image-image \\ (P@10)},
ymin=25,
ymax=46,
ytick={25, 35, 45},
tick label style={font=\tickfontsize},
yticklabel style={xshift=1ex}
]
\addplot[color=blue, mark=+, only marks] table[x=loss, y=a]{\alldata};
\end{axis}

&
\begin{axis}[
xlabel style={align=center},
xlabel={(d) Text-image \\ (P@10)},
ymax=61,
tick label style={font=\tickfontsize},
yticklabel style={xshift=1ex}
]
\addplot[color=green, mark=x, only marks] table[x=loss, y=b]{\alldata};
\end{axis}
\\
};

\end{tikzpicture}
% \shrinkcaptionmargin
% \halfshrinkcaptionmargin
\vspace{-1.5em}
\caption{
(a) shows pretraining validation loss at different epochs;
(b)-(d) shows correlation between the pretraining loss and the performance of three end tasks.
For (a) the x-axis shows the training epoch number, and for (b)-(d) the x-axis shows the negative value of the pretraining loss (i.e., $-\sL$) on a held-out validation set.
}
\label{fig:correlation}
\end{figure*}
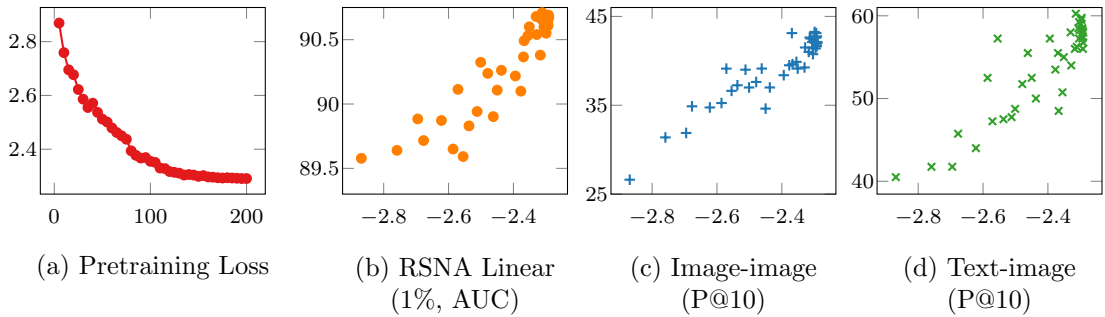

\paragraph{Correlation Between Contrastive Loss and End Task Performance.}
To understand the relation between a model's performance on the ConVIRT pretraining task and its performance on the downstream tasks, we ran an analysis where for every 5 epochs during the pretraining, we transferred the pretrained checkpoint to the downstream tasks and evaluate its performance.
The pretraining was run for a total of 200 epochs, and 40 points were obtained with varying validation loss and end task results.
\reffig{correlation} presents the results of the models' validation loss on the pretraining task, and its achieved performance on the RSNA 1\% data linear evaluation and the two retrieval tasks.
For all three tasks, we find a clear positive correlation between the pretraining performance and the end task performance.
This corroborates that by learning with the ConVIRT objectives, the image encoder learns gradually improved representations for the end tasks, and suggests that further improvement on the pretraining task may have positive impact on the end task performance.

\paragraph{Hyperparameter Analysis.}
We run experiments to study the impact of hyperparameters, and have the following observations.
First of all, similar to previous work on image-only contrastive learning \citep{chen2020simple,he2020momentum}, the pretraining results are most sensitive to the choice of the temperature value $\tau$. As shown in \reftab{hyperparam}, using a temperature much lower than the ideal value ($\tau = 0.01$) hurts the retrieval results, and a temperature much larger ($\tau = 1$) notably hurts the performance on all tasks.
Second, unlike previous work, changing batch size does not lead to substantial change in the classification results. At last, replacing the non-linear projection heads in $g_v$ and $g_u$ with linear layers hurts the retrieval results moderately, suggesting worse representations.
However, this is again not reflected notably in the RSNA classification results.

% hyperparameter analysis results
\begin{table*}
\begin{center}
% \small
\setlength{\tabcolsep}{1em}
\begin{tabular}{lccccc}
\toprule
& RSNA Linear & Image-Image & Text-Image \\
Settings  & (1\%, AUC) & (Prec@10) & (Prec@10) \\
\midrule
ConVIRT (default) & 90.7 & 42.9 & 57.5\\
\midrule
$\tau=0.01$ & 90.7 & 40.5 & 21.0 \\
$\tau=1$ & 89.6 & 25.0 & 31.0 \\
\midrule
bs $$= 16$$ & 90.3 & 40.0 & 55.8 \\
bs $$= 128$$ & 90.3 & 39.3 & 50.3 \\
\midrule
linear proj. & 90.6 & 40.8 & 55.8\\
\bottomrule
\end{tabular}
\end{center}
\caption{
Evaluation results with different hyperparameters, for the RSNA 1\% data linear evaluation, image-image retrieval and text-image retrieval tasks.
\emph{bs} represents batch size and \emph{linear proj.} represents using linear projection layers for $g_v$ and $g_u$.
Our default model uses $\tau=0.1$, bs $= 32$ and non-linear projections.
}
\label{tab:hyperparam}
\end{table*}

\paragraph{Limitations.}
This work mainly focuses on comparing ConVIRT against conventional ImageNet initialization, image captioning-based initialization, and image-only contrastive learning approaches including SimCLR and MoCo to demonstrate the data efficiency and effectiveness of image-text pretraining.
We did not compare our method against relevant subsequent studies that extended ConVIRT, such as LoVT \cite{muller2021joint} or GloRIA \cite{huang2021gloria}, mainly because such comparisons are included in these studies.

\section{Conclusion}

We presented ConVIRT, an unsupervised method for learning medical visual representations from paired descriptive text.
Our method relies on contrasting the image representations with the paired descriptive text via a bidirectional objective between the two modalities.
On 4 medical image classification tasks and 2 image retrieval tasks, ConVIRT outperformed other strong in-domain initialization methods, and led to representations with notably higher quality.
Compared to ImageNet pretraining, ConVIRT is able to achieve the same level of classification accuracy with an order of magnitude less labeled data. 
This is especially critical for the healthcare domain where data sparsity is an important issue, and the innovative cross-modality pretraining in ConVIRT is extensible to consider other modalities of data in this domain.
% In healthcare domain, ConVIRT is data efficient, which makes it easy to adapt to multiple medical image tasks because there are usually limited data for training machine learning systems. 
We thus hope that ConVIRT continues inspiring future work that makes more efficient use of multi-modal data for medical image understanding.

% ACKNOWLEDGEMENTS ONLY GO IN THE CAMERA-READY, NOT THE SUBMISSION
% \acks{Many thanks to all collaborators and funders!}

\bibliography{main}

\begin{thebibliography}{56}
\providecommand{\natexlab}[1]{#1}
\providecommand{\url}[1]{\texttt{#1}}
\expandafter\ifx\csname urlstyle\endcsname\relax
  \providecommand{\doi}[1]{doi: #1}\else
  \providecommand{\doi}{doi: \begingroup \urlstyle{rm}\Url}\fi

\bibitem[Abr{\`a}moff et~al.(2016)Abr{\`a}moff, Lou, Erginay, Clarida, Amelon,
  Folk, and Niemeijer]{abramoff2016improved}
Michael~David Abr{\`a}moff, Yiyue Lou, Ali Erginay, Warren Clarida, Ryan
  Amelon, James~C Folk, and Meindert Niemeijer.
\newblock Improved automated detection of diabetic retinopathy on a publicly
  available dataset through integration of deep learning.
\newblock \emph{Investigative Ophthalmology \& Visual Science}, 57\penalty0
  (13):\penalty0 5200--5206, 2016.

\bibitem[Alsentzer et~al.(2019)Alsentzer, Murphy, Boag, Weng, Jindi, Naumann,
  and McDermott]{alsentzer2019publicly}
Emily Alsentzer, John Murphy, William Boag, Wei-Hung Weng, Di~Jindi, Tristan
  Naumann, and Matthew McDermott.
\newblock Publicly available clinical {BERT} embeddings.
\newblock In \emph{Proceedings of the 2nd Clinical Natural Language Processing
  Workshop}, 2019.

\bibitem[Azizi et~al.(2021)Azizi, Mustafa, Ryan, Beaver, Freyberg, Deaton, Loh,
  Karthikesalingam, Kornblith, Chen, et~al.]{azizi2021big}
Shekoofeh Azizi, Basil Mustafa, Fiona Ryan, Zachary Beaver, Jan Freyberg,
  Jonathan Deaton, Aaron Loh, Alan Karthikesalingam, Simon Kornblith, Ting
  Chen, et~al.
\newblock Big self-supervised models advance medical image classification.
\newblock In \emph{Proceedings of the IEEE/CVF International Conference on
  Computer Vision (ICCV)}, 2021.

\bibitem[Chen et~al.(2020{\natexlab{a}})Chen, Kornblith, Norouzi, and
  Hinton]{chen2020simple}
Ting Chen, Simon Kornblith, Mohammad Norouzi, and Geoffrey Hinton.
\newblock A simple framework for contrastive learning of visual
  representations.
\newblock In \emph{International Conference on Machine Learning (ICML)},
  2020{\natexlab{a}}.

\bibitem[Chen et~al.(2020{\natexlab{b}})Chen, Fan, Girshick, and
  He]{chen2020improved}
Xinlei Chen, Haoqi Fan, Ross Girshick, and Kaiming He.
\newblock Improved baselines with momentum contrastive learning.
\newblock \emph{arXiv preprint arXiv:2003.04297}, 2020{\natexlab{b}}.

\bibitem[Cornia et~al.(2020)Cornia, Stefanini, Baraldi, and
  Cucchiara]{cornia2020meshed}
Marcella Cornia, Matteo Stefanini, Lorenzo Baraldi, and Rita Cucchiara.
\newblock Meshed-memory {Transformer} for image captioning.
\newblock In \emph{Proceedings of the IEEE/CVF Conference on Computer Vision
  and Pattern Recognition (CVPR)}, 2020.

\bibitem[De~Fauw et~al.(2018)De~Fauw, Ledsam, Romera-Paredes, Nikolov, Tomasev,
  Blackwell, Askham, Glorot, O’Donoghue, Visentin, et~al.]{de2018clinically}
Jeffrey De~Fauw, Joseph~R Ledsam, Bernardino Romera-Paredes, Stanislav Nikolov,
  Nenad Tomasev, Sam Blackwell, Harry Askham, Xavier Glorot, Brendan
  O’Donoghue, Daniel Visentin, et~al.
\newblock Clinically applicable deep learning for diagnosis and referral in
  retinal disease.
\newblock \emph{Nature Medicine}, 24\penalty0 (9):\penalty0 1342--1350, 2018.

\bibitem[Demner-Fushman et~al.(2016)Demner-Fushman, Kohli, Rosenman, Shooshan,
  Rodriguez, Antani, Thoma, and McDonald]{demner2016preparing}
Dina Demner-Fushman, Marc~D Kohli, Marc~B Rosenman, Sonya~E Shooshan, Laritza
  Rodriguez, Sameer Antani, George~R Thoma, and Clement~J McDonald.
\newblock Preparing a collection of radiology examinations for distribution and
  retrieval.
\newblock \emph{Journal of the American Medical Informatics Association},
  23\penalty0 (2):\penalty0 304--310, 2016.

\bibitem[Desai and Johnson(2021)]{desai2021virtex}
Karan Desai and Justin Johnson.
\newblock {VirTex}: Learning visual representations from textual annotations.
\newblock In \emph{Proceedings of the IEEE Conference on Computer Vision and
  Pattern Recognition (CVPR)}, 2021.

\bibitem[Devlin et~al.(2019)Devlin, Chang, Lee, and Toutanova]{devlin2019bert}
Jacob Devlin, Ming-Wei Chang, Kenton Lee, and Kristina Toutanova.
\newblock {BERT}: Pre-training of deep bidirectional transformers for language
  understanding.
\newblock In \emph{Proceedings of the 2019 Conference of the North American
  Chapter of the Association for Computational Linguistics: Human Language
  Technologies (NAACL-HLT)}, 2019.

\bibitem[Eslami et~al.(2021)Eslami, de~Melo, and Meinel]{eslami2021does}
Sedigheh Eslami, Gerard de~Melo, and Christoph Meinel.
\newblock Does {CLIP} benefit visual question answering in the medical domain
  as much as it does in the general domain?
\newblock \emph{arXiv preprint arXiv:2112.13906}, 2021.

\bibitem[Esteva et~al.(2017)Esteva, Kuprel, Novoa, Ko, Swetter, Blau, and
  Thrun]{esteva2017dermatologist}
Andre Esteva, Brett Kuprel, Roberto~A Novoa, Justin Ko, Susan~M Swetter,
  Helen~M Blau, and Sebastian Thrun.
\newblock Dermatologist-level classification of skin cancer with deep neural
  networks.
\newblock \emph{Nature}, 542\penalty0 (7639):\penalty0 115--118, 2017.

\bibitem[Grill et~al.(2020)Grill, Strub, Altch\'{e}, Tallec, Richemond,
  Buchatskaya, Doersch, Avila~Pires, Guo, Gheshlaghi~Azar, Piot, Kavukcuoglu,
  Munos, and Valko]{grill2020bootstrap}
Jean-Bastien Grill, Florian Strub, Florent Altch\'{e}, Corentin Tallec, Pierre
  Richemond, Elena Buchatskaya, Carl Doersch, Bernardo Avila~Pires, Zhaohan
  Guo, Mohammad Gheshlaghi~Azar, Bilal Piot, Koray Kavukcuoglu, Remi Munos, and
  Michal Valko.
\newblock Bootstrap your own latent: A new approach to self-supervised
  learning.
\newblock In \emph{Advances in Neural Information Processing Systems}, 2020.

\bibitem[Gulshan et~al.(2016)Gulshan, Peng, Coram, Stumpe, Wu, Narayanaswamy,
  Venugopalan, Widner, Madams, Cuadros, et~al.]{gulshan2016retinal}
Varun Gulshan, Lily Peng, Marc Coram, Martin~C Stumpe, Derek Wu, Arunachalam
  Narayanaswamy, Subhashini Venugopalan, Kasumi Widner, Tom Madams, Jorge
  Cuadros, et~al.
\newblock Development and validation of a deep learning algorithm for detection
  of diabetic retinopathy in retinal fundus photographs.
\newblock \emph{JAMA}, 316\penalty0 (22):\penalty0 2402--2410, 2016.

\bibitem[Gupta et~al.(2020)Gupta, Vahdat, Chechik, Yang, Kautz, and
  Hoiem]{gupta2020phrase}
Tanmay Gupta, Arash Vahdat, Gal Chechik, Xiaodong Yang, Jan Kautz, and Derek
  Hoiem.
\newblock Contrastive learning for weakly supervised phrase grounding.
\newblock In \emph{Proceedings of the 16th European Conference on Computer
  Vision (ECCV)}, 2020.

\bibitem[Han et~al.(2021)Han, Chen, Tewfik, Ding, and Peng]{han2021pneumonia}
Yan Han, Chongyan Chen, Ahmed Tewfik, Ying Ding, and Yifan Peng.
\newblock Pneumonia detection on chest x-ray using radiomic features and
  contrastive learning.
\newblock In \emph{2021 IEEE 18th International Symposium on Biomedical Imaging
  (ISBI)}. IEEE, 2021.

\bibitem[He et~al.(2016)He, Zhang, Ren, and Sun]{he2016deep}
Kaiming He, Xiangyu Zhang, Shaoqing Ren, and Jian Sun.
\newblock Deep residual learning for image recognition.
\newblock In \emph{Proceedings of the IEEE Conference on Computer Vision and
  Pattern Recognition (CVPR)}, 2016.

\bibitem[He et~al.(2020)He, Fan, Wu, Xie, and Girshick]{he2020momentum}
Kaiming He, Haoqi Fan, Yuxin Wu, Saining Xie, and Ross Girshick.
\newblock Momentum contrast for unsupervised visual representation learning.
\newblock In \emph{Proceedings of the IEEE/CVF Conference on Computer Vision
  and Pattern Recognition (CVPR)}, 2020.

\bibitem[Heiliger et~al.(2022)Heiliger, Sekuboyina, Menze, Egger, and
  Kleesiek]{heiliger2022beyond}
Lars Heiliger, Anjany Sekuboyina, Bjoern Menze, Jan Egger, and Jens Kleesiek.
\newblock Beyond medical imaging: A review of multimodal deep learning in
  radiology.
\newblock \emph{TechRxiv preprint}, 2022.

\bibitem[H{\'e}naff et~al.(2020)H{\'e}naff, Srinivas, De~Fauw, Razavi, Doersch,
  Eslami, and Oord]{henaff2020cpc}
Olivier~J H{\'e}naff, Aravind Srinivas, Jeffrey De~Fauw, Ali Razavi, Carl
  Doersch, SM~Eslami, and Aaron van~den Oord.
\newblock Data-efficient image recognition with contrastive predictive coding.
\newblock In \emph{International Conference on Machine Learning (ICML)}, 2020.

\bibitem[Huang et~al.(2021)Huang, Shen, Lungren, and Yeung]{huang2021gloria}
Shih-Cheng Huang, Liyue Shen, Matthew~P Lungren, and Serena Yeung.
\newblock {GLoRIA}: A multimodal global-local representation learning framework
  for label-efficient medical image recognition.
\newblock In \emph{Proceedings of the IEEE/CVF International Conference on
  Computer Vision (ICCV)}, 2021.

\bibitem[Ilharco et~al.(2021)Ilharco, Zellers, Farhadi, and
  Hajishirzi]{ilharco2020probing}
Gabriel Ilharco, Rowan Zellers, Ali Farhadi, and Hannaneh Hajishirzi.
\newblock Probing contextual language models for common ground with visual
  representations.
\newblock In \emph{Proceedings of the 2021 Conference of the North American
  Chapter of the Association for Computational Linguistics: Human Language
  Technologies (NAACL-HLT)}, 2021.

\bibitem[Irvin et~al.(2019)Irvin, Rajpurkar, Ko, Yu, Ciurea-Ilcus, Chute,
  Marklund, Haghgoo, Ball, Shpanskaya, et~al.]{irvin2019chexpert}
Jeremy Irvin, Pranav Rajpurkar, Michael Ko, Yifan Yu, Silviana Ciurea-Ilcus,
  Chris Chute, Henrik Marklund, Behzad Haghgoo, Robyn Ball, Katie Shpanskaya,
  et~al.
\newblock {CheXpert}: A large chest radiograph dataset with uncertainty labels
  and expert comparison.
\newblock In \emph{Proceedings of the AAAI Conference on Artificial
  Intelligence}, 2019.

\bibitem[Jia et~al.(2021)Jia, Yang, Xia, Chen, Parekh, Pham, Le, Sung, Li, and
  Duerig]{jia2021scaling}
Chao Jia, Yinfei Yang, Ye~Xia, Yi-Ting Chen, Zarana Parekh, Hieu Pham, Quoc Le,
  Yun-Hsuan Sung, Zhen Li, and Tom Duerig.
\newblock Scaling up visual and vision-language representation learning with
  noisy text supervision.
\newblock In \emph{Proceedings of the 38th International Conference on Machine
  Learning}, 2021.

\bibitem[Jing et~al.(2018)Jing, Xie, and Xing]{jing2018automatic}
Baoyu Jing, Pengtao Xie, and Eric Xing.
\newblock On the automatic generation of medical imaging reports.
\newblock In \emph{Proceedings of the 56th Annual Meeting of the Association
  for Computational Linguistics (ACL)}, 2018.

\bibitem[Johnson et~al.(2019)Johnson, Pollard, Berkowitz, Greenbaum, Lungren,
  Deng, Mark, and Horng]{johnson2019mimic}
Alistair~EW Johnson, Tom~J Pollard, Seth~J Berkowitz, Nathaniel~R Greenbaum,
  Matthew~P Lungren, Chih-ying Deng, Roger~G Mark, and Steven Horng.
\newblock {MIMIC-CXR}, a de-identified publicly available database of chest
  radiographs with free-text reports.
\newblock \emph{Scientific Data}, 6, 2019.

\bibitem[Kingma and Ba(2015)]{kingma2014adam}
Diederik~P Kingma and Jimmy Ba.
\newblock Adam: A method for stochastic optimization.
\newblock In \emph{The 2015 International Conference for Learning
  Representations}, 2015.

\bibitem[Liao et~al.(2021)Liao, Moyer, Cha, Quigley, Berkowitz, Horng, Golland,
  and Wells]{liao2021multimodal}
Ruizhi Liao, Daniel Moyer, Miriam Cha, Keegan Quigley, Seth Berkowitz, Steven
  Horng, Polina Golland, and William~M Wells.
\newblock Multimodal representation learning via maximization of local mutual
  information.
\newblock In \emph{International Conference on Medical Image Computing and
  Computer-Assisted Intervention}, 2021.

\bibitem[Lin et~al.(2014)Lin, Maire, Belongie, Hays, Perona, Ramanan,
  Doll{\'a}r, and Zitnick]{lin2014microsoft}
Tsung-Yi Lin, Michael Maire, Serge Belongie, James Hays, Pietro Perona, Deva
  Ramanan, Piotr Doll{\'a}r, and C~Lawrence Zitnick.
\newblock Microsoft {COCO}: Common objects in context.
\newblock In \emph{European Conference on Computer Vision (ECCV)}, 2014.

\bibitem[Liu et~al.(2019)Liu, Hsu, McDermott, Boag, Weng, Szolovits, and
  Ghassemi]{liu2019clinically}
Guanxiong Liu, Tzu-Ming~Harry Hsu, Matthew McDermott, Willie Boag, Wei-Hung
  Weng, Peter Szolovits, and Marzyeh Ghassemi.
\newblock Clinically accurate chest {X}-ray report generation.
\newblock In \emph{Machine Learning for Healthcare Conference}, 2019.

\bibitem[Lu et~al.(2019)Lu, Batra, Parikh, and Lee]{lu2019vilbert}
Jiasen Lu, Dhruv Batra, Devi Parikh, and Stefan Lee.
\newblock {ViLBERT}: Pretraining task-agnostic visiolinguistic representations
  for vision-and-language tasks.
\newblock In \emph{Advances in Neural Information Processing Systems}, 2019.

\bibitem[Maaten and Hinton(2008)]{maaten2008visualizing}
Laurens van~der Maaten and Geoffrey Hinton.
\newblock Visualizing data using {t-SNE}.
\newblock \emph{Journal of Machine Learning Research}, 9\penalty0
  (Nov):\penalty0 2579--2605, 2008.

\bibitem[Manning et~al.(2014)Manning, Surdeanu, Bauer, Finkel, Bethard, and
  McClosky]{manning2014stanford}
Christopher~D. Manning, Mihai Surdeanu, John Bauer, Jenny Finkel, Steven~J.
  Bethard, and David McClosky.
\newblock The {Stanford} {CoreNLP} natural language processing toolkit.
\newblock In \emph{Association for Computational Linguistics (ACL) System
  Demonstrations}, 2014.

\bibitem[Miura et~al.(2021)Miura, Zhang, Langlotz, and
  Jurafsky]{miura2020improving}
Yasuhide Miura, Yuhao Zhang, Curtis~P. Langlotz, and Dan Jurafsky.
\newblock Improving factual completeness and consistency of image-to-text
  radiology report generation.
\newblock In \emph{Proceedings of the 2021 Conference of the North American
  Chapter of the Association for Computational Linguistics: Human Language
  Technologies (NAACL-HLT)}, 2021.

\bibitem[M{\"u}ller et~al.(2021)M{\"u}ller, Kaissis, Zou, and
  R{\"u}ckert]{muller2021joint}
Philip M{\"u}ller, Georgios Kaissis, Congyu Zou, and Daniel R{\"u}ckert.
\newblock Joint learning of localized representations from medical images and
  reports.
\newblock \emph{arXiv preprint arXiv:2112.02889}, 2021.

\bibitem[Oord et~al.(2018)Oord, Li, and Vinyals]{oord2018representation}
Aaron van~den Oord, Yazhe Li, and Oriol Vinyals.
\newblock Representation learning with contrastive predictive coding.
\newblock \emph{arXiv preprint arXiv:1807.03748}, 2018.

\bibitem[Radford et~al.(2021)Radford, Kim, Hallacy, Ramesh, Goh, Agarwal,
  Sastry, Askell, Mishkin, Clark, et~al.]{radford2021learning}
Alec Radford, Jong~Wook Kim, Chris Hallacy, Aditya Ramesh, Gabriel Goh,
  Sandhini Agarwal, Girish Sastry, Amanda Askell, Pamela Mishkin, Jack Clark,
  et~al.
\newblock Learning transferable visual models from natural language
  supervision.
\newblock In \emph{International Conference on Machine Learning}, 2021.

\bibitem[Raghu et~al.(2019)Raghu, Zhang, Kleinberg, and
  Bengio]{raghu2019transfusion}
Maithra Raghu, Chiyuan Zhang, Jon Kleinberg, and Samy Bengio.
\newblock Transfusion: Understanding transfer learning for medical imaging.
\newblock In \emph{Advances in Neural Information Processing Systems}, 2019.

\bibitem[Rajpurkar et~al.(2018{\natexlab{a}})Rajpurkar, Irvin, Bagul, Ding,
  Duan, Mehta, Yang, Zhu, Laird, Ball, et~al.]{rajpurkar2017mura}
Pranav Rajpurkar, Jeremy Irvin, Aarti Bagul, Daisy Ding, Tony Duan, Hershel
  Mehta, Brandon Yang, Kaylie Zhu, Dillon Laird, Robyn~L Ball, et~al.
\newblock {MURA}: Large dataset for abnormality detection in musculoskeletal
  radiographs.
\newblock In \emph{1st Conference on Medical Imaging with Deep Learning
  (MIDL)}, 2018{\natexlab{a}}.

\bibitem[Rajpurkar et~al.(2018{\natexlab{b}})Rajpurkar, Irvin, Ball, Zhu, Yang,
  Mehta, Duan, Ding, Bagul, Langlotz, et~al.]{rajpurkar2018chexnext}
Pranav Rajpurkar, Jeremy Irvin, Robyn~L Ball, Kaylie Zhu, Brandon Yang, Hershel
  Mehta, Tony Duan, Daisy Ding, Aarti Bagul, Curtis~P Langlotz, et~al.
\newblock Deep learning for chest radiograph diagnosis: A retrospective
  comparison of the {CheXNeXt} algorithm to practicing radiologists.
\newblock \emph{PLoS Medicine}, 15\penalty0 (11):\penalty0 e1002686,
  2018{\natexlab{b}}.

\bibitem[Russakovsky et~al.(2015)Russakovsky, Deng, Su, Krause, Satheesh, Ma,
  Huang, Karpathy, Khosla, Bernstein, et~al.]{russakovsky2015imagenet}
Olga Russakovsky, Jia Deng, Hao Su, Jonathan Krause, Sanjeev Satheesh, Sean Ma,
  Zhiheng Huang, Andrej Karpathy, Aditya Khosla, Michael Bernstein, et~al.
\newblock {ImageNet} large scale visual recognition challenge.
\newblock \emph{International Journal of Computer Vision}, 115\penalty0
  (3):\penalty0 211--252, 2015.

\bibitem[Sariyildiz et~al.(2020)Sariyildiz, Perez, and
  Larlus]{sariyildiz2020learning}
Mert~Bulent Sariyildiz, Julien Perez, and Diane Larlus.
\newblock Learning visual representations with caption annotations.
\newblock In \emph{Proceedings of the 16th European Conference on Computer
  Vision (ECCV)}, 2020.

\bibitem[Shih et~al.(2019)Shih, Wu, Halabi, Kohli, Prevedello, Cook, Sharma,
  Amorosa, Arteaga, Galperin-Aizenberg, et~al.]{shih2019augmenting}
George Shih, Carol~C Wu, Safwan~S Halabi, Marc~D Kohli, Luciano~M Prevedello,
  Tessa~S Cook, Arjun Sharma, Judith~K Amorosa, Veronica Arteaga, Maya
  Galperin-Aizenberg, et~al.
\newblock Augmenting the {N}ational {I}nstitutes of {H}ealth chest radiograph
  dataset with expert annotations of possible pneumonia.
\newblock \emph{Radiology: Artificial Intelligence}, 1\penalty0 (1):\penalty0
  e180041, 2019.

\bibitem[Simonyan et~al.(2014)Simonyan, Vedaldi, and
  Zisserman]{simonyan2014saliency}
Karen Simonyan, Andrea Vedaldi, and Andrew Zisserman.
\newblock Deep inside convolutional networks: Visualising image classification
  models and saliency maps.
\newblock In \emph{ICLR Workshop}, 2014.

\bibitem[Sowrirajan et~al.(2021)Sowrirajan, Yang, Ng, and
  Rajpurkar]{sowrirajan2021moco}
Hari Sowrirajan, Jingbo Yang, Andrew~Y Ng, and Pranav Rajpurkar.
\newblock {MoCo} pretraining improves representation and transferability of
  chest {X}-ray models.
\newblock In \emph{Medical Imaging with Deep Learning}, pages 728--744. PMLR,
  2021.

\bibitem[Su et~al.(2020)Su, Zhu, Cao, Li, Lu, Wei, and Dai]{su2019vlbert}
Weijie Su, Xizhou Zhu, Yue Cao, Bin Li, Lewei Lu, Furu Wei, and Jifeng Dai.
\newblock {VL-BERT}: Pre-training of generic visual-linguistic representations.
\newblock In \emph{International Conference on Learning Representations
  (ICLR)}, 2020.

\bibitem[Tan and Bansal(2019)]{tan2019lxmert}
Hao Tan and Mohit Bansal.
\newblock {LXMERT}: Learning cross-modality encoder representations from
  transformers.
\newblock In \emph{Proceedings of the 2019 Conference on Empirical Methods in
  Natural Language Processing and the 9th International Joint Conference on
  Natural Language Processing (EMNLP-IJCNLP)}, 2019.

\bibitem[Vedantam et~al.(2015)Vedantam, Lawrence~Zitnick, and
  Parikh]{vedantam2015cider}
Ramakrishna Vedantam, C~Lawrence~Zitnick, and Devi Parikh.
\newblock {CIDEr}: Consensus-based image description evaluation.
\newblock In \emph{Proceedings of the IEEE Conference on Computer Vision and
  Pattern Recognition (CVPR)}, 2015.

\bibitem[Vu et~al.(2021)Vu, Wang, Balachandar, Liu, Ng, and
  Rajpurkar]{vu2021medaug}
Yen Nhi~Truong Vu, Richard Wang, Niranjan Balachandar, Can Liu, Andrew~Y Ng,
  and Pranav Rajpurkar.
\newblock {MedAug}: Contrastive learning leveraging patient metadata improves
  representations for chest x-ray interpretation.
\newblock In \emph{Machine Learning for Healthcare Conference}, 2021.

\bibitem[Wang and Wong(2020)]{wang2020covid}
Linda Wang and Alexander Wong.
\newblock {COVID-Net}: A tailored deep convolutional neural network design for
  detection of {COVID-19} cases from chest {X}-ray images.
\newblock \emph{arXiv preprint arXiv:2003.09871}, 2020.

\bibitem[Wang et~al.(2017)Wang, Peng, Lu, Lu, Bagheri, and
  Summers]{wang2017chestx}
Xiaosong Wang, Yifan Peng, Le~Lu, Zhiyong Lu, Mohammadhadi Bagheri, and
  Ronald~M Summers.
\newblock {ChestX-ray8}: Hospital-scale chest {X}-ray database and benchmarks
  on weakly-supervised classification and localization of common thorax
  diseases.
\newblock In \emph{Proceedings of the IEEE Conference on Computer Vision and
  Pattern Recognition (CVPR)}, 2017.

\bibitem[Wang et~al.(2018)Wang, Peng, Lu, Lu, and Summers]{wang2018tienet}
Xiaosong Wang, Yifan Peng, Le~Lu, Zhiyong Lu, and Ronald~M Summers.
\newblock {TieNet}: Text-image embedding network for common thorax disease
  classification and reporting in chest {X}-rays.
\newblock In \emph{Proceedings of the IEEE Conference on Computer Vision and
  Pattern Recognition (CVPR)}, 2018.

\bibitem[Wang et~al.(2021)Wang, Xu, Tam, Yang, and Xu]{wang2021self}
Xiaosong Wang, Ziyue Xu, Leo Tam, Dong Yang, and Daguang Xu.
\newblock Self-supervised image-text pre-training with mixed data in chest
  x-rays.
\newblock \emph{arXiv preprint arXiv:2103.16022}, 2021.

\bibitem[Wolf et~al.(2020)Wolf, Debut, Sanh, Chaumond, Delangue, Moi, Cistac,
  Rault, Louf, Funtowicz, Davison, Shleifer, von Platen, Ma, Jernite, Plu, Xu,
  Le~Scao, Gugger, Drame, Lhoest, and Rush]{wolf2019huggingfaces}
Thomas Wolf, Lysandre Debut, Victor Sanh, Julien Chaumond, Clement Delangue,
  Anthony Moi, Pierric Cistac, Tim Rault, Remi Louf, Morgan Funtowicz, Joe
  Davison, Sam Shleifer, Patrick von Platen, Clara Ma, Yacine Jernite, Julien
  Plu, Canwen Xu, Teven Le~Scao, Sylvain Gugger, Mariama Drame, Quentin Lhoest,
  and Alexander Rush.
\newblock Transformers: State-of-the-art natural language processing.
\newblock In \emph{Proceedings of the 2020 Conference on Empirical Methods in
  Natural Language Processing (EMNLP): System Demonstrations}, 2020.

\bibitem[Xu et~al.(2015)Xu, Ba, Kiros, Cho, Courville, Salakhudinov, Zemel, and
  Bengio]{xu2015show}
Kelvin Xu, Jimmy Ba, Ryan Kiros, Kyunghyun Cho, Aaron Courville, Ruslan
  Salakhudinov, Rich Zemel, and Yoshua Bengio.
\newblock Show, attend and tell: Neural image caption generation with visual
  attention.
\newblock In \emph{International Conference on Machine Learning (ICML)}, 2015.

\bibitem[Zang and Wang(2021)]{zang2021scehr}
Chengxi Zang and Fei Wang.
\newblock Scehr: Supervised contrastive learning for clinical risk prediction
  using electronic health records.
\newblock \emph{arXiv preprint arXiv:2110.04943}, 2021.

\end{thebibliography}

\appendix
\section{Model Implementation and Pretraining Details}
\label{sec:pretraining-details}

\paragraph{Dataset Preprocessing.}
For the MIMIC-CXR chest radiograph dataset, we use the publicly available JPG version of it.%
\footnote{\url{https://physionet.org/content/mimic-cxr-jpg/2.0.0/}}
For both the MIMIC-CXR chest dataset and the Rhode Island Hospital bone image datasets, we resize the image files to have a size of 256 on the larger side.
For the textual radiology report data, we first tokenize all reports with the default English tokenizer in version 4.0.0 of the CoreNLP library \citep{manning2014stanford}.
Next, we keep only the \emph{Findings} and \emph{Impression} sections and remove all other sections.
We remove all image-text pairings from the dataset where the text section is empty or has less than 3 tokens.
This preprocessing procedure gives us about 217k total image-text pairs for pretraining our chest image encoder and 48k total pairs for pretraining our bone image encoder.

\paragraph{Image and Text Encoders.}
For the image encoder, we use the standard ResNet50 implementation provided by the torchvision library.
For the text encoder, we use the BERT base encoder offered by the Transformers library \citep{wolf2019huggingfaces} and initialize it with the ClinicalBERT model \citep{alsentzer2019publicly} pretrained on the MIMIC clinical notes.
We also experimented with training a specialized BERT encoder on a large collection of radiology notes but found that it made no substantial difference in the pretraining results.
At pretraining time we freeze the embeddings and the first 6 layers of this BERT encoder, and only fine-tune the last 6 layers for our contrastive task.

\paragraph{Other Hyperparameters.}
For contrastive learning, we use projection layers with an output dimension $d = 512$, a temperature value $\tau = 0.1$, a loss weight $\lambda = 0.75$.
These hyperparameter settings are obtained by comparing the linear evaluation validation scores on the RSNA image classification task with the pretrained ResNet50 weights.
For the image transformation family $\sT$, we adopt the implementations offered by the torchvision library.%
\footnote{\url{https://github.com/pytorch/vision}}
We apply \emph{random cropping} with a ratio sampled from $[0.6, 1.0]$;
\emph{horizontal flipping} with $p = 0.5$;
\emph{affine transformation} with a degree sampled from $[-20, 20]$, max horizontal and vertical translation fractions of 0.1, and a scaling factor sampled from $[0.95, 1.05]$;
\emph{color jittering} with brightness and contrast adjustment ratios sampled from $[0.6, 1.4]$;
and \emph{Gaussian blur} with $\sigma \in [0.1, 3.0]$.
All images are resized to $224{\times}224$ after the transformation $t_v$ is applied.
Limited by computational resources, we arrive at these image transformation parameters via preliminary experiments rather than a systematic search.

\paragraph{Pretraining Details.}
At pretraining time, for each dataset, we randomly sample 5k image-text pairs to form a held-out validation set.
We we use the Adam optimizer \citep{kingma2014adam} with an initial learning rate of 1e-4 and weight decay of 1e-6.
We initialize the image encoder with ImageNet pretrained weights at the beginning of pretraining, and use a fixed batch size of 32.
We calculate the validation loss every 5000 steps, and if the validation loss does not decrease after 5 straight evaluation runs, we anneal the learning rate by a factor of 0.5.
We stop pretraining after 200 evaluation runs, and save the model checkpoint that achieves the lowest validation loss.
For efficiency, we employ mixed-precision training, and for reference, the whole pretraining run on the MIMIC-CXR dataset took about 3 days on a single Titan RTX GPU card.

\section{Image Classification Experiments}
\label{sec:classification-details}

We prepared and used the 4 image classification datasets following the procedures below:

\begin{enumerate}[leftmargin=*,align=left]
\item \textbf{RSNA Pneumonia Detection} \citep{wang2017chestx,shih2019augmenting}: 
we used the original version of this dataset available at its Kaggle page,%
\footnote{\url{https://www.kaggle.com/c/rsna-pneumonia-detection-challenge}}
which contains 25184/1500/3000 annotated images in its training/validation/test sets, respectively.

\item \textbf{CheXpert} image classification \citep{irvin2019chexpert}:
we downloaded the original version of this dataset from its official website.%
\footnote{\url{https://stanfordmlgroup.github.io/competitions/chexpert/}}
Since the original expert-labeled test set of this dataset is hidden and not included as part of the release, we instead followed \citet{raghu2019transfusion} and used the original expert-labeled validation set as our test set, and randomly sampled 5000 images from the original training set for validation purpose.
The resulting dataset contains 218414/5000/234 images in each split.

\item \textbf{COVIDx} image classification \citep{wang2020covid}:
we prepared this dataset following the scripts provided by its authors.%
\footnote{\url{https://github.com/lindawangg/COVID-Net}}
We used the version 4 of this dataset, the latest version at the time of this work.
We additionally randomly sampled 300 images from the training set for validation, resulting in a dataset with 13598/300/300 images in each split.

\item \textbf{MURA} bony abnormality detection \citep{rajpurkar2017mura}:
we downloaded the original version of this dataset from its website.%
\footnote{\url{https://stanfordmlgroup.github.io/competitions/mura/}}
Similar to the CheXpert dataset, we again used the original validation set as our test set, and randomly sampled 10\% images from the training set for validation, resulting in a dataset with 33078/3730/3197 images in each split.
Different from the other 3 datasets, the MURA dataset uses patient-level evaluation, meaning that the prediction results from different images of the same patient needs to be aggregated to produce a final prediction for the patient, which is then scored against the gold patient label.
We therefore followed \citet{rajpurkar2017mura} and at test time aggregated result for a patient by averaging the predicted probabilities from multiple images.
\end{enumerate}

\paragraph{Classification Model Training Details.}
For all models that require ImageNet pretrained initialization, we use the pretrained weights from torchvision, which achieves an ImageNet top-5 error rate of 7.13\%.
For all datasets, we first zero-pad the input image to be square, and then resize it to be $224{\times}224$.
For training, we use the Adam optimizer with an initial learning rate of 1e-3 for the COVIDx task and 1e-4 for the other three tasks.
We additionally apply a weight decay of 1e-6 and a dropout before the last classification layer with $p=0.2$ in all tasks.
All classification models are trained with a batch size of 64.
In the fine-tuning evaluation setting, we first ``warmup'' the classification head by freezing the CNN weights and only training the classification head with a learning rate of 1e-3 for 200 steps, after which we unfreeze the CNN weights and fine-tune the entire network together.
Validation score is obtained after each epoch of training and we anneal the learning rate by a factor of 0.5 if the validation score is not improved after 3 epochs.
The training is stopped after no validation improvement is observed for 10 straight epochs, at which point the model checkpoint with the highest validation score is evaluated on the test set.

% example textual queries
\begin{table*}[h]
\begin{center}
\begin{tabular}{ll}
\toprule
Image Category & Example Textual Query \\
\midrule
Atelectasis & Platelike opacity likely represents atelectasis.\\
Cardiomegaly & The cardiac silhouette is enlarged.\\
Edema & The presence of hazy opacity suggests interstitial pulmonary edema.\\
Fracture & A cortical step off indicates the presence of a fracture.\\
Pleural Effusion & The pleural space is partially filled with fluid.\\
Pneumonia & A pulmonary opacity with ill defined borders likely represents pneumonia.\\
Pneumothorax & A medial pneumothorax is present adjacent to the heart.\\
No Finding & No clinically significant radiographic abnormalities.\\
\bottomrule
\end{tabular}
\end{center}
% \shrinkcaptionmargin
\caption{Example textual queries for each of the 8 categories in the text-image retrieval task.}
\label{tab:text-query}
\end{table*}

\section{Image-image Retrieval Dataset Collection}
\label{sec:image-image}

We create the CheXpert $8{\times}200$ Retrieval Dataset with 8 different abnormality categories commonly found in Chest radiograph images, including \emph{atelectasis}, \emph{cardiomegaly}, \emph{edema}, \emph{fracture}, \emph{pleural effusion}, \emph{pneumonia}, \emph{pneumothorax} and a special \emph{no finding} category indicating that no obvious abnormality is found in the image.
We create the dataset by reusing existing rule-labeled annotations in the CheXpert dataset \citep{irvin2019chexpert} and additional expert annotations.
To create the candidate images for a category label $\ell$, we go through all images in the CheXpert training set, and keep an image as a candidate image if only its label for $\ell$ is positive and all other categories negative.
We only include images with this ``exclusive positivity'' as candidate images, mainly to avoid confounding results between categories in retrieval evaluation.

To create the query images for a category $\ell$, we again first pre-select 50 exclusively positive images for this category in the CheXpert training set (with all candidate images excluded).
Next, we ask a board-certified radiologist to examine each of the 50 images, and exclude images that: 1) might indicate additional abnormalities other than $\ell$, 2) have uncommon color or contrast distortions in the image, or 3) are not well posed during the capture of the image.
This procedure is mainly to avoid including query images that have uncommon features and may therefore bias the retrieval evaluation results.
At the end, we aggregate the annotation results from the radiologist and keep 10 query images for each abnormality category.

\section{Text-image Retrieval Dataset Collection}
\label{sec:text-image}

For the text-image retrieval dataset, we first reuse all candidate images from the CheXpert $8{\times}200$ image-image retrieval dataset described above, with 200 images for each of 8 categories.
To create the textual queries for each abnormality category, we ask a board-certified radiologist to write at least 5 different sentences that he will use to describe this abnormality in radiology reporting.
We additionally set the following requirements:
1) the sentences must describe the category with no ambiguity and must not include other categories;
2) the sentences must be diverse from each other;
and 3) the sentences should not include very specific anatomic locations or rare clinical observations.
At the end, we aggregate the results and keep 5 textual queries for each abnormality category.
For reference, we present example textual queries in \reftab{text-query}.

\section{Experiments on Image-Only Contrastive Learning Methods}
\label{sec:image-only}

We run experiments with two popular image-only contrastive visual representation learning methods: SimCLR \citep{chen2020simple} and MoCo v2 \citep{chen2020improved}.
For a fair comparison, in both experiments we use the exact same set of images from the MIMIC-CXR dataset that we use in the pretraining of our method and the baselines.
Our settings for each method are:
\begin{itemize}[leftmargin=*,align=left]
\item \textbf{SimCLR}: We use the open PyTorch implementation available at \url{https://github.com/sthalles/SimCLR}.
For image encoder we use ResNet50.
We use cosine similarity in the loss function, set the temperature value to 0.1 and set the output dimension to 128.
We use the default image augmentation functions in the paper except for the \emph{color jittering} transformation where we set the saturation and hue adjustment to 0 due to the monochrome nature of our medical images.
For training, we use the Adam optimizer with an initial learning rate of 3e-4 and weight decay of 1e-4.
We set batch size to 128 and run training on a single GPU card for 100 epochs, as we find that increasing the batch size or number of epochs does not lead to improved results.
We use the default settings for all other parameters.

\item \textbf{MoCo v2}: We use the authors' PyTorch implementation available at \url{https://github.com/facebookresearch/moco}.
For image encoder we use ResNet50.
We follow the default MoCo v2 setting and use a temperature value of 0.07 and an output dimension of 128.
Similarly, we adopt the default image augmentation functions except for the \emph{color jittering} transformation where we set the saturation and hue adjustment to 0.
For training, we use the SGD optimizer with a learning rate of 0.0075 and weight decay of 1e-4.
We use a batch size of 64 and a queue size of 4096, and run parallel training on two GPU cards for 100 epochs, as we find that further increasing the batch size or number of epochs does not lead to improved results.
During training, we anneal the learning rate by a factor of 0.1 at the 60th and 80th epochs.
\end{itemize}

% \section{Hyperparameter Analysis}
% \label{sec:hyperparam}

% Similar to previous work on unsupervised image representation learning \citep{chen2020simple, he2020momentum}, we first find that the effectiveness of ConVIRT pretraining is most sensitive to the temperature value $\tau$.
% As shown in \reftab{hyperparam}, using a temperature much lower than the ideal value ($\tau = 0.01$) hurts the retrieval results, and a temperature much larger ($\tau = 1$) notably hurts the performance on all tasks.
% Unlike previous work, we find that using a smaller or larger batch size hurts the retrieval performance, but neither setup brings substantial impact to the classification results.
% Lastly, we find that replacing the non-linear projection heads in $g_v$ and $g_u$ with linear layers hurts the retrieval results moderately, suggesting worse representations.
% However, this is again not reflected notably in the RSNA classification results.

\end{document}